\documentclass[lettersize,journal]{IEEEtran}
\usepackage{amsmath,amsfonts}
\usepackage{algorithmic}
\usepackage{algorithm}
\usepackage{array}
\usepackage{enumitem}
\usepackage[caption=false,font=normalsize,labelfont=sf,textfont=sf]{subfig}
\usepackage{textcomp}
\usepackage{stfloats}
\usepackage{url}
\usepackage{multirow}
\usepackage{verbatim}
\usepackage{graphicx}
\usepackage{cite}
\usepackage{utfsym}
\usepackage{makecell}
\usepackage{color, xcolor}
\usepackage{hyperref}

\hypersetup{
	colorlinks = true,
	linkcolor=blue,
	filecolor=blue,
	urlcolor=blue,
	citecolor=blue,
}

\newcommand\copyrighttext{%
		\footnotesize \textcopyright 2025 IEEE. Personal use of this material is permitted.  Permission from IEEE must be obtained for all other uses, in any current or future media, including reprinting/republishing this material for advertising or promotional purposes, creating new collective works, for resale or redistribution to servers or lists, or reuse of any copyrighted component of this work in other works.
		DOI: \href{https://ieeexplore.ieee.org/document/10848213}{10.1109/TVCG.2025.3532081}}
\newcommand\copyrightnotice{%
		\begin{tikzpicture}[remember picture,overlay]
				\node[anchor=north,yshift=-30pt] at (current page.north) {\fbox{\parbox{\dimexpr\textwidth-\fboxsep-\fboxrule\relax}{\copyrighttext}}};
			\end{tikzpicture}%
	}

\begin{document}
	\title{Aligning Instance-Semantic Sparse Representation towards Unsupervised Object Segmentation and Shape Abstraction with Repeatable Primitives}
	
	\author{Jiaxin Li, Hongxing Wang, \IEEEmembership{Member, IEEE}, Jiawei Tan, Zhilong Ou, and Junsong Yuan, \IEEEmembership{Fellow, IEEE}
		\thanks{This work is supported in part by the National Natural Science Foundation of China under Grant 61976029 and the Key Project of Chongqing Technology Innovation and Application Development under Grant cstc2021jscx-gksbX0033. Jiaxin Li, Hongxing Wang, Jiawei Tan, and Zhilong Ou are with the School of Big Data and Software Engineering, Chongqing University, Chongqing 401331, China. Junsong Yuan is with the Department of Computer Science and Engineering, State University of New York at Buffalo, NY 14260 USA (e-mail: jiaxin\_li@cqu.edu.cn; ihxwang@cqu.edu.cn; jwtan@cqu.edu.cn; zlou@stu.cqu.edu.cn; jsyuan@buffalo.edu) Corresponding author: Hongxing Wang.}
	}
	
	\markboth{IEEE Transactions on Visualization and Computer Graphics}%
	{Shell \MakeLowercase{\textit{et al.}}: A Sample Article Using IEEEtran.cls for IEEE Journals}

	\maketitle
	\copyrightnotice
	
	\begin{abstract}
		Understanding 3D object shapes necessitates shape representation by object parts abstracted from results of instance and semantic segmentation. Promising shape representations enable computers to interpret a shape with meaningful parts and identify their repeatability. However, supervised shape representations depend on costly annotation efforts, while current unsupervised methods work under strong semantic priors and involve multi-stage training, thereby limiting their generalization and deployment in shape reasoning and understanding. Driven by the tendency of high-dimensional semantically similar features to lie in or near low-dimensional subspaces, we introduce a one-stage, fully unsupervised framework towards semantic-aware shape representation. This framework produces joint instance segmentation, semantic segmentation, and shape abstraction through sparse representation and feature alignment of object parts in a high-dimensional space. For sparse representation, we devise a sparse latent membership pursuit method that models each object part feature as a sparse convex combination of point features at either the semantic or instance level, promoting part features in the same subspace to exhibit similar semantics. For feature alignment, we customize an attention-based strategy in the feature space to align instance- and semantic-level object part features and reconstruct the input shape using both of them, ensuring geometric reusability and semantic consistency of object parts. To firm up semantic disambiguation, we construct cascade unfrozen learning on geometric parameters of object parts. Experiments conducted on benchmark datasets confirm that our approach results in instance- and semantic-level joint segmentation and shape abstraction with repeatable primitives, providing coherent semantic interpretations of 3D object shapes across categories in a one-stage, fully unsupervised manner, without relying on annotations or heuristic semantic priors. Code will be released at \url{https://github.com/L-Jiaxin/AISSR}.
	\end{abstract}
	
	\begin{IEEEkeywords}
		Point cloud, semantic representation, unsupervised learning, shape abstraction, segmentation.
	\end{IEEEkeywords}

\section{Introduction}
\IEEEPARstart{S}{emantics}, integral to human perception and reasoning, pertains to the meaning of entities \cite{lundgard2021accessible}. Specifically within 3D object shape representation, semantic shape representation extends beyond the geometries of object parts, which are the sole focus of instance shape representations, to encompass their semantic meanings. The holistic semantic representation can promote understanding, interpretation, and even generation of objects, benefiting various fields such as augmented reality, autonomous driving, and game development \cite{chang2015shapenet, ioannidou2017deep, hu2018semantic, tesema2023point}. Recently, significant strides have been made in weakly supervised and unsupervised semantic representation for 3D object point clouds to relieve the annotation requirements \cite{guo2021deep, xiao2023unsupervised}. Notably, both representations require mappings between predicted one-hot labels and predefined semantic concepts \cite{chen2019bae}. The current state-of-the-art approaches typically require separate models per object category to incorporate strong heuristic information and rely on inefficient multi-stage training, which pose deployment challenges. To the best of our knowledge, purely unsupervised, one-stage, category-free learning-based methods for semantic object shape representations remain unexplored.

\begin{figure}[!tp]
	\centering
	\includegraphics[width=3.5in]{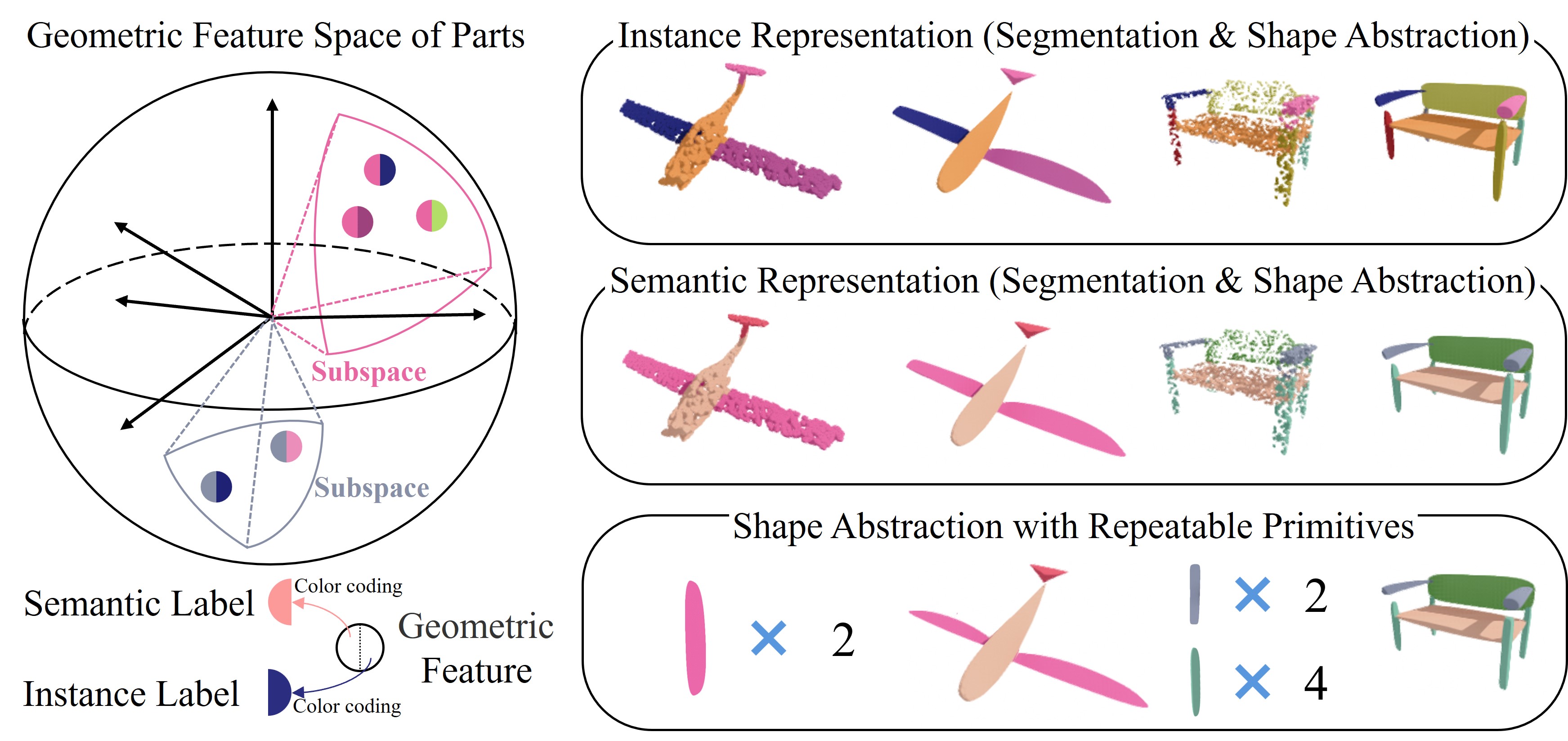}
	\caption{Unsupervised semantic shape representation via feature alignment in low-dimensional subspaces of a high-dimensional space (sparse representation): instance-level / semantic-level segmentation and shape abstraction w/o or w/ repeatable primitive-reconstructed parts. Each circle represents the geometric feature of an object part.}
	\label{fig: keypoint}
\end{figure}

In this study, we embark on the first step towards achieving a purely unsupervised, one-stage, category-free learning-based 3D semantic representation of unoriented man-made object point clouds, which unites five representations: ({\romannumeral1}) instance shape abstraction, employing primitives for succinct and key geometric representation \cite{biederman1987recognition}, ({\romannumeral2}) instance segmentation, defining geometries of individual parts through grouped points, ({\romannumeral3}) semantic shape abstraction, prioritizing the semantics conveyed by primitives, ({\romannumeral4}) semantic segmentation, emphasizing the semantics of grouped points, and ({\romannumeral5}) shape abstraction with repeatable primitives, involving primitive reuse \cite{deng2022unsupervised}, as illustrated in Fig. \ref{fig: keypoint}. Contrastly, the pioneering method \cite{shu2016unsupervised} introduces a three-stage algorithm: extracting high-dimensional features from over-segmented mesh patches using non-learning descriptors, reducing dimensionality through autoencoder (AE), and clustering the features. Similarly, BAE-NET \cite{chen2019bae}, supervised by ground truth (GT) occupancy functions, uses a branched AE to learn compact part features by reducing dimensionality. This approach aggregates semantically consistent parts into a single field, like four chair legs as one field. In contrast to feature compactness, ProGRIP \cite{deng2022unsupervised} predicts $M$ cuboids, each characterized by 6 poses, and utilizes the Hungarian algorithm \cite{kuhn1955hungarian} to enforce geometric compactness. It leverages cuboids predicted by CAS \cite{yang2021unsupervised} as GT for initial training stage, and subsequently refines predicted cuboids using GT occupancy functions. Conversely, the latest method \cite{umam2024unsupervised} uses powerful priors (limited semantics, \emph{e.g.}, 3 for chairs and 2 for tables) to enforce clustering in superpoint features generated by its first-staged network. 

\begin{figure*}[!tp]
	\centering
	\includegraphics[width=\textwidth]{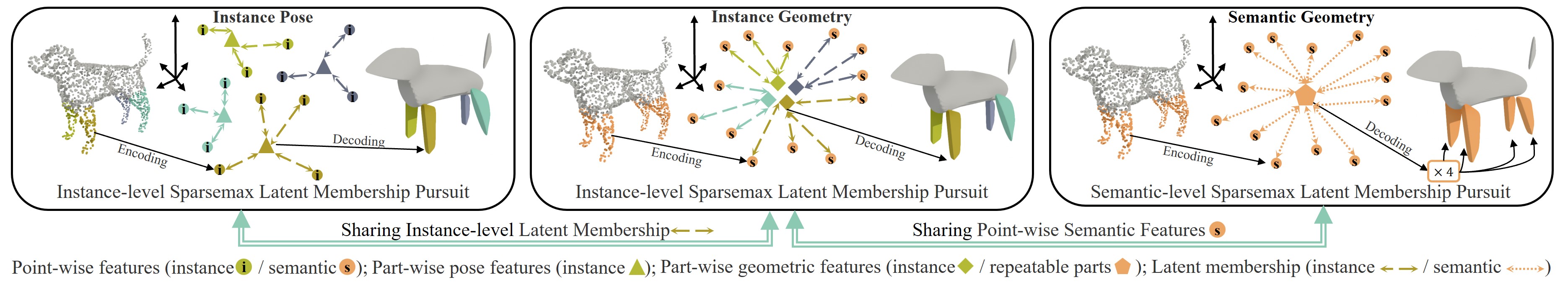}
	\caption{
		Alignment of instance-level geometric features with semantic-level geometric features. Instance-level latent membership pursuit determines the pose and geometry of each individual part, while semantic-level latent membership pursuit captures shared geometry among repeatable parts. Instance-level geometric features introduce variations among the four legs (middle), while semantic-level geometric features ensure consistent and repeatable geometry (right). Note that shape abstraction retains a few defining geometric characteristics, rather than the pursuit of exhaustive shape reconstruction.
	}
	\label{fig: membership}
\end{figure*}

Drawing on the idea that compact features in high-dimensional spaces enhance semantic representations \cite{shu2016unsupervised, chen2019bae}, our approach leverages sparse representation and feature alignment to improve unsupervised semantic representations. Sparse representation of features, as highlighted by \cite{wright2010sparse}, reveals semantics since high-dimensional features with similar semantics tend to lie in or near low-dimensional subspaces \cite{su2015multi, mu2017representing}. To construct compact part features without imposing specific constraints, we introduce Sparsemax \cite{martins2016softmax} into latent membership pursuit (LMP) \cite{li2024shared}, abbreviated as SLMP, yielding: ({\romannumeral1}) a sparse weight matrix to transform compact part features into sparse linear expressions of point features, and ({\romannumeral2}) a probability assignment matrix for segmentation. Consequently, we develop instance-level SLMP (I-SLMP) to extract instance-level geometric features for each object part, and semantic-level SLMP (S-SLMP) to derive semantic-level geometric features for each semantic concept. In fact, semantic-level geometric features should align with their semantically consistent instance-level counterparts within the same subspaces to consistently describe geometries across repeatable object parts, as shown in Fig. \ref{fig: keypoint}. and Fig. \ref{fig: membership}. To achieve this alignment and strengthen intra-subspace similarity, we reconstruct the same object shape using both instance-level geometric features and semantic-level geometric features, which discovers instance-semantic affinity from the view of feature alignment \cite{baltruvsaitis2018multimodal, zhang2023rethinking}.

\begin{figure}[!tp]
	\centering
	\includegraphics[width=3.5in]{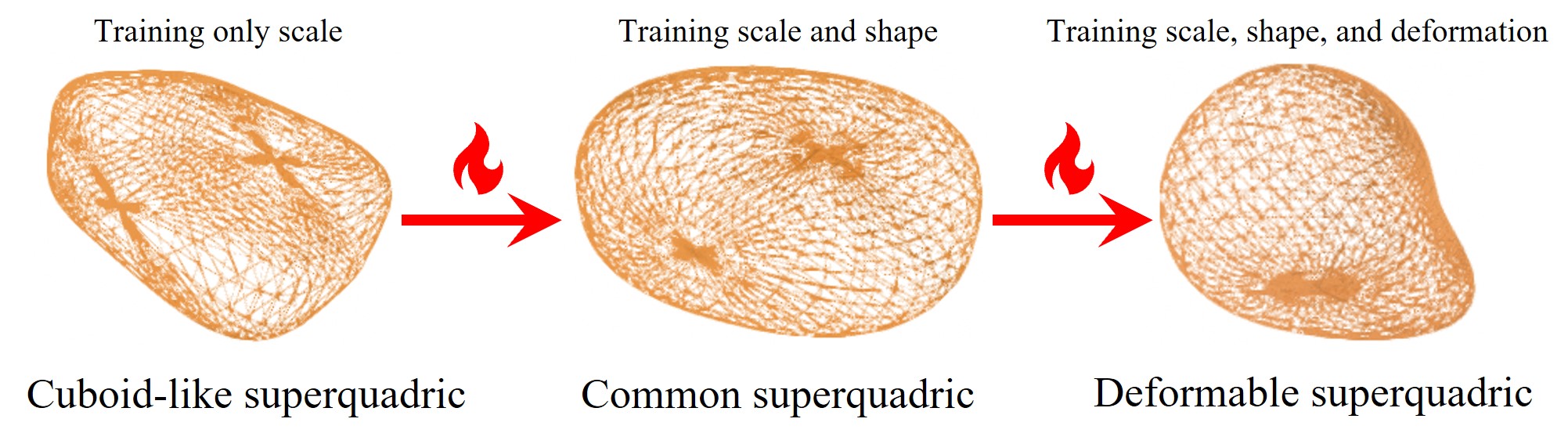}
	\caption{Cascade unfreezing learning on DSQ parameters.}
	\label{fig: unfreezing}
\end{figure}

Apart from sparsity and alignment, the one-to-one correspondence between primitive parameters and geometry holds significant importance for semantic-level joint shape abstraction and segmentation. Deformable superquadrics (DSQs) \cite{barr1981superquadrics, barr1984global, biederman1987recognition} are particularly pivotal in achieving visually pleasing shape abstraction, offering distinct advantages over alternative primitives such as lines \cite{qin2019mass, lin2022seg, zhang2022point, qu2024sketch2human}, spheres \cite{thiery2013sphere, yang2020p2mat}, cubes \cite{zou20173d, tulsiani2017learning, li2017grass, sun2019learning, yang2021unsupervised}, implicit functions \cite{fougerolle2005boolean, genova2019learning, genova2020local}, mapped geons \cite{wang2018adaptive, paschalidou2021neural}, multi-geons \cite{li2019supervised}, and superquadrics \cite{paschalidou2019superquadrics, paschalidou2020learning, liu2022robust, wu2022primitive, kim2022dsqnet, li2022editvae}. DSQs excel in accurately capturing diverse geometric variations, thus facilitating joint improvements in shape abstraction and segmentation through LMP. However, the inherent multisolution of DSQs-based shape reconstruction \cite{solina1990recovery} can counteract the alignment and lead to semantic ambiguity. To maximize the unique correspondence between semantics and primitive geometry, we constrain the geometric parameter space of DSQs \cite{solina1990recovery} to ensure unique solutions and employ a cascade unfreezing learning on our DSQ geometric parameter decoder to enhance robust semantic identification, as depicted in Fig. \ref{fig: unfreezing}. In contrast to prior work, our model, trained across categories, achieves a fully unsupervised, one-stage, learning-based 3D semantic representation through the combined use of sparsity and alignment.

To sum up, our main contributions are:
\begin{itemize}
	\item To the best of our knowledge, we are the first to propose a fully unsupervised, one-stage, category-free learning-based 3D semantic representation of unoriented man-made object point clouds.
	\item We introduce the Sparsemax function into the latent membership pursuit for sparse representations of part geometric features to unveil semantic information.
	\item We align semantic-level part features with instance-level part features in the same space for geometric reusability and semantic consistency.
	\item We employ a cascade unfreezing learning on deformable superquadric parameters to improve the robustness of unsupervised semantic representation.
\end{itemize}

\section{Related Work}
\label{sec: related work}
\subsection{Instance-level Shape Abstraction}
\label{sec: instance shape abstraction}
Instance-level shape abstraction is a methodology aimed at encapsulating key characteristics of individual object parts using primitives, notably implicit functions, cubes, and superquadrics. In the case of implicit functions, the methods \cite{genova2019learning, genova2020local, niu2022rim} employ 3D Gaussians to derive general shape templates, whereas alternative approaches \cite{sharma2018csgnet, kania2020ucsg} reconstruct shapes using constructive solid geometry trees. Regarding cube-based approaches, 3D-PRNN \cite{zou20173d} and a CNN-based unsupervised network \cite{tulsiani2017learning} predict cubes for shape abstractions. Furthermore, GRASS \cite{li2017grass} studies the structure of cubic primitives, CAS \cite{yang2021unsupervised} combines shape abstraction and segmentation, and DPF-Net \cite{shuai2023dpf} refines cube-based parts via GT occupancy field supervision. In the realm of superquadrics, SQA \cite{paschalidou2019superquadrics} and EditVAE \cite{li2022editvae} parses objects into superquadric representations. Non-learning methods \cite{liu2022robust, wu2022primitive} leverage probabilistic models to predict superquadrics. DSQNet \cite{kim2022dsqnet} constructs shape abstractions through deformable superquadrics for object grasping. LMP \cite{li2024shared} achieves joint shape abstraction and segmentation using deformable superquadrics via a single matrix.

Our method not only achieves instance-level 3D shape representation but also reveals semantic relationships between individual object parts, all without the need for any supervision. Furthermore, our approach involves the reuse of repeatable primitives.

\subsection{Semantic-level Shape Representation}
\label{sec: semantic shape representation}
Semantic-level shape representation refers to the high-level understanding and explanation of 3D objects, encompassing geometric, semantic, and structural information. For instance, \cite{zheng2010non, li2011globfit, digne2017sparse} exploit repeated geometries for point cloud denoising and completion. Prior to the advent of deep learning, early research defines semantic-level shape representation as co-segmentation \cite{shu2016unsupervised}. This approach, which involves segmenting instance-level parts within a single object guided by underlying semantic relevance, yields superior results compared to instance segmentation \cite{kalogerakis2010learning, sidi2011unsupervised}. The first deep learning-based 3D shape segmentation \cite{shu2016unsupervised} employs a three-stage algorithm: non-learning feature descriptors for over-segmented mesh patches, deep learning for high-level feature extraction, and clustering for instance-level representations. BAE-NET \cite{chen2019bae} embodies the Occam's Razor by finding the simplest part representations for a shape collection via a branch architecture. Supervised by GT occupancy functions, each branch predictor either gives an implicit field representing a combination of semantically similar parts or simply outputs nothing. ProGRIP \cite{deng2022unsupervised} builds upon the instance-level shape abstraction and segmentation produced by trained CAS \cite{yang2021unsupervised}. To achieve the reuse of subroutines for repeatable parts, ProGRIP predicts geometric and pose parameters of cuboids involved in one-to-many relationships. Subsequently, ProGRIP enhances shape reconstruction through GT occupancy functions. VoxAttention \cite{wu2023attentionbased} uses GT part transformation matrix parameters and GT occupancy functions to train its three-stage model for part assembly. A semi-supervised framework uses sparse scribble semantic labels as weakly supervisory information for 3D shape segmentation \cite{shu2024semi}. SGAS \cite{zhang2024point} is a four-stage process for point cloud part editing, achieving semantic feature disentanglement through supervision of GT semantic labels. The latest unsupervised method \cite{umam2024unsupervised} constructs the Superpoint Generation Network (SG-Net) and the Part Aggregation Network (PA-Net) through sequential training using category-specific datasets. However, PA-Net carefully predefines a small semantic number, forcing all points to conform to specified clusters.

Beyond the aforementioned methods, our approach stands out as a fully unsupervised, one-stage, learning-based method that eliminates the necessity for category-specific training or pre-training while supporting a larger semantic number.

\subsection{Unsupervised Point Cloud Representation	Learning}
\label{sec: unsupervised representation}
Unsupervised point cloud representation learning aims to extract robust and versatile features from unlabeled data, categorized into pre-training reliant and pre-training independent models. FoldingNet \cite{yang2018foldingnet} employs self-reconstruction pre-training and introduces a folding-based decoder for classification tasks, which transforms a 2D grid into a 3D mesh. Latent-GAN \cite{achlioptas2018learning} pioneers deep generative models for point clouds via reconstruction, focusing on 3D recognition and shape editing. PU-GAN \cite{li2019pu} is a GAN-based point cloud upsampling network trained through reconstruction. Without relying on pre-training, Growsp \cite{zhang2023growsp} presents the first purely unsupervised 3D semantic segmentation pipeline tailored for large-scale point clouds. PointDC \cite{chen2023pointdc} incorporates Cross-Modal Distillation (CMD) and Super-Voxel Clustering (SVC). CMD aggregates multi-view visual features to enhance point representation training, while SVC iteratively clusters point features to extract semantic classes. $\text{U3DS}^3$ \cite{liu2024u3ds3} generates superpoints based on geometry for clustering, followed by iterative training using clustering-based pseudo-labels. Moreover, $\text{U3DS}^3$ \cite{liu2024u3ds3} utilizes voxelized features for representation learning.

Our approach does not require pre-training or iterative training. As far as we are aware, it represents the initial implementation of purely unsupervised and one-stage learning-based methods targeting small-scale point clouds for semantic 3D shape representations.

\section{Method}
\label{sec: method}
\begin{figure*}[!tp]
	\centering
	\includegraphics[width=\linewidth]{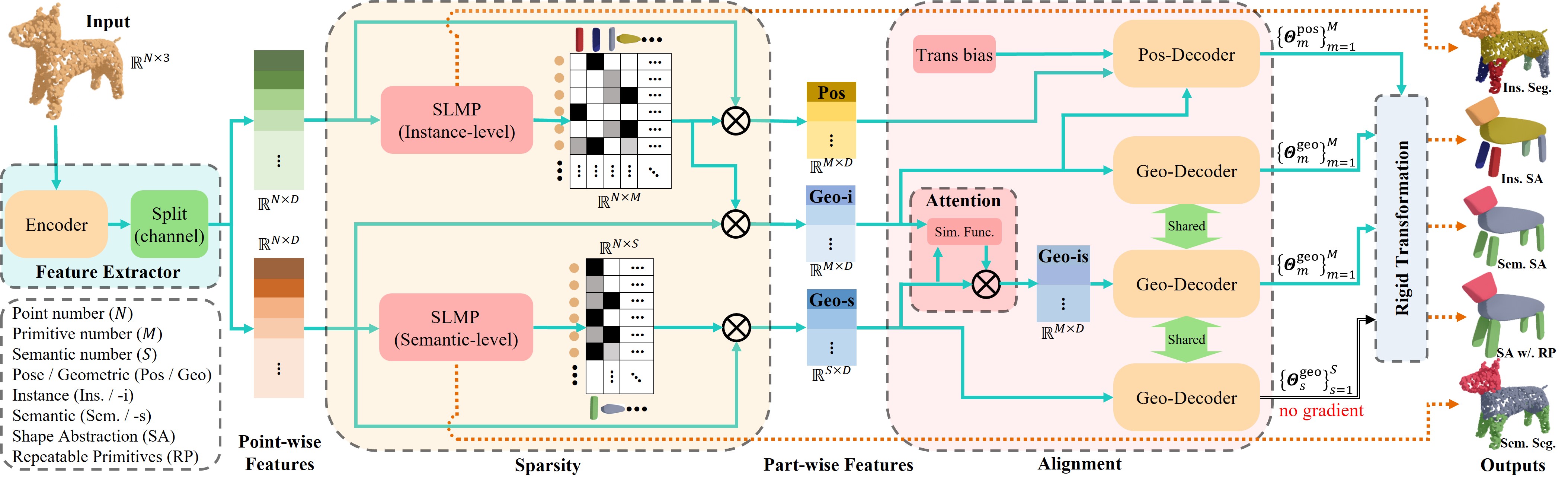}
	\caption{Overview of our pipeline. $\bigotimes$ represents matrix multiplication. SLMP outputs segmentation and the sparse weight matrix to construct part features in low-dimensional subspaces, refer to Fig. \ref{fig: sparseLMP} and Section \ref{sec: sparse latent membership pursuit}. Trans bias, Pos-Decoder, and Geo-Decoder are used to predict primitive parameters, refer to Fig. \ref{fig: decoder} for details. Sim Func indicates the inner product similarity with an adaptive temperature parameter, defined by (\ref{eq: attention alignment with temperature}) and (\ref{eq: tau}). Part-wise features include: Instance-level pose features (Pos), Instance-level geometric features (Geo-i), Semantic-level geometric features (Geo-s), and semantic-aligned instance-level geometric features (Geo-is). Outputs, from top to bottom, are: instance segmentation (\emph{cf.} Instance-level SLMP), instance shape abstraction (\emph{cf.} Geo-i), semantic shape abstraction (\emph{cf.} Geo-is), semantic shape abstraction with repeatable parts (\emph{cf.} Geo-s), and semantic segmentation (\emph{cf.} Semantic-level SLMP).
	}
	\label{fig: architecture}
\end{figure*}

We propose an autoencoder-structured framework to implement our one-stage, fully unsupervised approach for semantic shape representation, enabling semantic-level joint segmentation and shape abstraction with repeatable primitives, as depicted in Fig. \ref{fig: architecture}.

\subsection{Preliminaries}
\label{sec: preliminaries}
\subsubsection{Deformable superquadric}
\label{sec: preliminaries dsq}
As shown in Fig. \ref{fig: unfreezing}, superquadrics (SQs) \cite{barr1981superquadrics} are recognized for their versatile geometric representations achieved through a minimal set of parameters \cite{schultz2010superquadric}, enabling them to model shapes like ellipsoids, octahedra, cylinders, \emph{etc}. In this paper, superquadrics specifically denotes superellipsoids. Deformable superquadrics (DSQs) \cite{barr1984global, biederman1987recognition} integrate global tapering and bending on superquadrics. The depiction of a DSQ in our shape abstraction comprises two components: geometric parameters $\boldsymbol{\theta}^\text{geo}$ and pose parameters $\boldsymbol{\theta}^\text{pos}$. For geometric parameters, DSQs are defined by size parameter $ \boldsymbol{a} \in \mathbb{R}^{3} $, shape parameter $ \boldsymbol{\epsilon} \in \mathbb{R}^{2} $, tapering coefficient $ \boldsymbol{k} \in \mathbb{R}^{2} $, bending direction angle $\alpha \in \mathbb{R} $, and curvature parameter $b \in \mathbb{R}$. To mitigate potential semantic ambiguities arising from different parameters corresponding to the same geometry, we limit the shape parameters to the interval $\left[ 0.1, 1 \right ]$, as suggested in \cite{solina1990recovery}. Additionally, to ensure numerical stability, we restrict the DSQ parameter space as outlined in Table \ref{tab: parameters}. Regarding pose parameters, we use $\boldsymbol{t} \in \mathbb{R}^3 $ to denote translation and $\boldsymbol{r} \in \mathbb{R}^4$ (a quaternion representation) to give rotation. We define the transformation order of DSQs as tapering, bending, rotation, and translation, following the conventional geometric transformation order \cite{barr1984global, biederman1987recognition}. Thus, we systematically describe the parameters of a DSQ as $ \boldsymbol{\theta}:= \left\{ \boldsymbol{\theta}^\text{geo}, \boldsymbol{\theta}^\text{pos} \right\} = \left\{ \boldsymbol{a}, \boldsymbol{\epsilon}, \boldsymbol{k}, b, \alpha, \boldsymbol{t}, \boldsymbol{r} \right\} \in \mathbb{R}^{16}$. 

\begin{table}[h]
	\renewcommand{\arraystretch}{1.2}
	\begin{center}
		\caption{Range of Parameters.}
		\label{tab: parameters}
		\begin{tabular}{|cc|c|c|c|c|}
			\hline
				\multicolumn{2}{|c|}{Parameters} & Symbol & Dimensions & Constraints \\ \hline
			\multicolumn{1}{|c|}{\multirow{5}{*}{\makecell[c]{geometry\\$\left(\boldsymbol{\theta}^\text{geo}\right)$}}}
			& size & $\boldsymbol{a}$ & $\mathbb{R}^{3}$ & $\left[0.02, 0.82\right]$ \\ \cline{2-5}
			\multicolumn{1}{|c|}{} 
			& shape & $\boldsymbol{\epsilon}$ & $\mathbb{R}^{2}$ & $\left[0.2, 1\right]$ \\ \cline{2-5}
			\multicolumn{1}{|c|}{} 
			& tapering & $\boldsymbol{k}$ & $\mathbb{R}^{2}$ & $\left[-0.9, 0.9\right]$ \\ \cline{2-5}
			\multicolumn{1}{|c|}{} 
			& \multirow{2}{*}{bending} & $b$ & $\mathbb{R}^{1}$ & $\left[0.01, 0.75\right]$ \\ \cline{3-5}
			\multicolumn{1}{|c|}{} 
			& & $\alpha$ & $\mathbb{R}^{1}$ & $\left[-\pi/2, \pi/2\right]$ \\ \hline
			\multicolumn{1}{|c|}{\multirow{2}{*}{\makecell[c]{pose\\$\left(\boldsymbol{\theta}^\text{pos}\right)$}}} 
			& translation & $\boldsymbol{t}$ & $\mathbb{R}^{3}$ & $\left[-1, 1\right]$ \\ \cline{2-5}
			\multicolumn{1}{|c|}{} 
			& rotation & $\boldsymbol{r}$ & $\mathbb{R}^{4}$ & $\mathbb{S}^{3}$ \\ \hline
		\end{tabular}
	\end{center}
\end{table}

\subsubsection{Latent Membership Pursuit}
\label{sec: preliminaries lmp}
Latent Membership Pursuit (LMP) \cite{li2024shared} constructs a single matrix $\boldsymbol{J} \in \mathbb{R}^{M \times N}$ from point-wise features, referred to as membership matrix, to generate instance-level joint shape abstraction and segmentation from an unoriented point cloud $\boldsymbol{X}$ of an object containing $N$ points, where $M$ represents the maximum number of parts. For instance shape abstraction, LMP applies softmax operations along the rows of $\boldsymbol{J}$ to form a weight matrix $ \boldsymbol{W} \in \mathbb{R}^{N \times M}$ for transforming part features $\boldsymbol{F}^{\text{part}} \in \mathbb{R}^{D\times M}$ into linear expressions of point features $\boldsymbol{F}^{\text{pc}} \in \mathbb{R}^{D \times N}$ within a $D$-dimensional latent space. For instance segmentation, the assignment matrix $\boldsymbol{P} \in \mathbb{R}^{M \times N}$, whose element represents the probability that a point is assigned to an object part, is obtained through column-wise softmax applied to $\boldsymbol{J}$. LMP is defined by the following equations:
\begin{equation}
	\begin{cases} 
		\boldsymbol{W} = \sigma_{\text{col}} \left( \boldsymbol{J}^{\mathsf{T}} \right),
		\\ 
		\boldsymbol{P} = \sigma_{\text{col}} \left( \boldsymbol{J} \right),
		\\
		\boldsymbol{F}^{\text{part}} = \boldsymbol{F}^{\text{pc}} \boldsymbol{W}.
	\end{cases}
	\label{eq: model} 
\end{equation}
Here, $ \sigma_{\text{col}} \left(\cdot \right) $ signifies the softmax function applied along columns of the input matrix.

\subsection{Problem Formulation}
\label{sec: problem formulation}
Given an unoriented point cloud $\boldsymbol{X}$ of an object containing $N$ points, where $\boldsymbol{X}:=\left\{ \boldsymbol{x}_n \in \mathbb{R}^{3} \right\}_{n=1}^N $, our objective is to produce semantic-level joint segmentation and shape abstraction with repeatable primitives, including instance-level segmentation and shape abstraction. In the context of instance level, we designate $M$ as the maximum number of primitives (parts) per object. At the semantic level, $S$ represents the maximum number of semantics for a single object. In total, we produce five outputs:
\begin{enumerate}[label=(\roman*)]
	\item Instance segmentation, $\boldsymbol{P}^{\text{Ins}} \in \mathbb{R}^{M \times N}$.
	\item Semantic segmentation, $\boldsymbol{P}^{\text{Sem}} \in \mathbb{R}^{S \times N}$.
	\item Instance shape abstraction, $\boldsymbol{\Theta}^{\text{Ins}}:=\left\{\boldsymbol{\theta}^{\text{Ins}}_m \in \mathbb{R}^{16} \right\}_{m=1}^M$.
	\item Semantic shape abstraction, $\boldsymbol{\Theta}^{\text{Sem}}:=\left\{\boldsymbol{\theta}^{\text{Sem}}_m \in \mathbb{R}^{16} \right\}_{m=1}^M$.
	\item Semantic shape abstraction with repeatable primitives (parts), $\boldsymbol{\Theta}^{\text{Rep}}:=\left\{\boldsymbol{\theta}^{\text{Rep}}_m \in \mathbb{R}^{16} \right\}_{m=1}^M$.
\end{enumerate}

\subsection{Encoder}
We employ the segmentation framework of PointNet++ \cite{qi2017pointnet, qi2017pointnet++} to encode an unoriented point cloud $\boldsymbol{X}$ into a $(2D)$-dimensional latent space, yielding the point features $ \boldsymbol{F}^{\text{pc}} \in \mathbb{R}^{2D \times N}$. PointNet++ is an efficient and widely used point cloud feature extractor, which is agnostic to specific tasks and aligns well with our objectives. Our goal is to discern the semantics of parts reconstructed by primitives, which necessitates obtaining instance features and semantic features. To achieve this, we equally split the point features along the channel dimension into instance pursuit point features $ \boldsymbol{F}^{\text{Ins}} \in \mathbb{R}^{D \times N}$ and semantic pursuit point features $ \boldsymbol{F}^{\text{Sem}} \in \mathbb{R}^{D \times N}$, as shown in Fig. \ref{fig: architecture}.

\begin{figure}[!tbp]
	\centering
	\includegraphics[width=\linewidth]{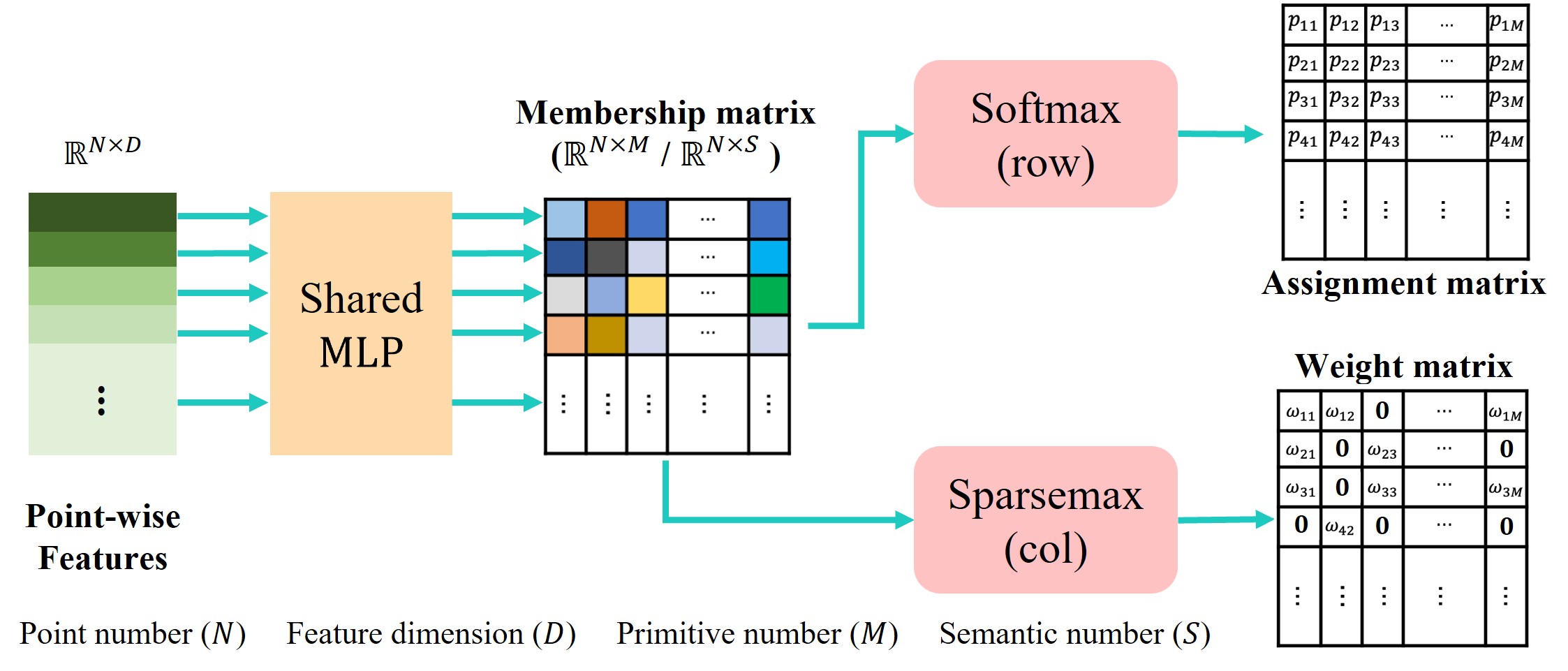}
	\caption{Sparse Latent Membership Pursuit (SLMP). The matrices in the figure have been transposed for presentational convenience.}
	\label{fig: sparseLMP}
\end{figure}

\subsection{Sparsity Induced Latent Membership Pursuit}
\label{sec: sparse latent membership pursuit}
Motivated by the observation that features with similar semantics often exhibit a degenerate structure in high-dimensional spaces and lie in or near low-dimensional subspaces \cite{wright2010sparse}, we introduce Sparsemax \cite{martins2016softmax} into LMP (\emph{cf.} Section \ref{sec: preliminaries lmp}) for sparse representation of object part geometric features, naming it Sparse Latent Membership Pursuit (SLMP), which unveils semantic information. To clarify, given any vector $\boldsymbol{z} \in \mathbb{R}^C$, the definitions of Softmax and Sparsemax are as follows:
\begin{align}
		\operatorname{Softmax}_i \left( \boldsymbol{z}\right) &= \frac{\exp(z_i)}{\sum_{j=1}^{C}{\exp(z_j)}},\\
		\operatorname{Sparsemax} \left( \boldsymbol{z}\right) &:= \mathop{\arg\min}\limits_{\boldsymbol{g} \in \Delta ^ {C - 1}} \Vert \boldsymbol{g} - \boldsymbol{z} \Vert^2,
	\label{eq: softmax sparsemax} 
\end{align}
where $\Delta ^ {C - 1} := \left\{ \boldsymbol{g} \in \mathbb{R}^C \mid \boldsymbol{1}^\mathsf{T}\boldsymbol{g}=1, \forall g_i \ge 0 \right\}$ is the $\left(C-1\right)$-dimensional simplex. The defining characteristic of Sparsemax lies in its capability to generate sparse posterior distributions by assigning zero probability to trivial variables. This characteristic makes Sparsemax well-suited to discern pertinent point-part memberships within LMP for part reconstruction, thus producing a sparse representation of part features that enhances semantic interpretability. Additionally, Sparsemax exhibits simplicity in evaluation and cost-effectiveness for differentiation \cite{martins2016softmax}. Our Sparse Latent Membership Pursuit is defined as
\begin{equation}
	\begin{cases} 
		\boldsymbol{W} = \operatorname{Sparsemax}_{\text{col}} \left( \boldsymbol{J}^{\mathsf{T}} \right),
		\\ 
		\boldsymbol{P} = \operatorname{Softmax}_{\text{col}} \left( \boldsymbol{J} \right),
		\\
		\boldsymbol{F}^{\text{part}} = \boldsymbol{F}^{\text{pc}} \boldsymbol{W}.
	\end{cases}
	\label{eq: slmp} 
\end{equation}

Importantly, our SLMP constructs the assignment matrix $\boldsymbol{P}$ for segmentation using Softmax, which maintains consistency with LMP. Note that both instance labels and semantic labels are determined by the highest probability elements in the columns of the corresponding assignment matrix. Hence, we employ Softmax to ensure that every membership entry in the membership matrix $\boldsymbol{J}$ can receive gradient updates.

An illustration of our SLMP implementation is shown in Fig. \ref{fig: sparseLMP}. To obtain the membership matrices for joint shape abstraction and segmentation, we employ two independent three-layer Shared MLPs \cite{qi2017pointnet} to construct the instance-level membership matrix $\boldsymbol{J}^{\text{Ins}} \in \mathbb{R}^{M \times N}$ and the semantic-level membership matrix $\boldsymbol{J}^{\text{Sem}} \in \mathbb{R}^{S \times N}$ from $\boldsymbol{F}^{\text{Ins}}$ and $\boldsymbol{F}^{\text{Sem}}$, respectively. Then, based on (\ref{eq: slmp}), we have
\begin{equation}
	\begin{cases} 
		\boldsymbol{W}^{\text{Ins}} = \operatorname{Sparsemax}_{\text{col}} \left(\left( \boldsymbol{{J}^{\text{Ins}}}\right)^{\mathsf{T}} \right),
		\\ 
		\boldsymbol{W}^{\text{Sem}} = \operatorname{Sparsemax}_{\text{col}} \left(\left( \boldsymbol{{J}^{\text{Sem}}}\right)^{\mathsf{T}} \right),
		\\ 
		\boldsymbol{P}^{\text{Ins}} = \operatorname{Softmax}_{\text{col}} \left( \boldsymbol{{J}^{\text{Ins}}} \right),
		\\ 
		\boldsymbol{P}^{\text{Sem}} = \operatorname{Softmax}_{\text{col}} \left( \boldsymbol{{J}^{\text{Sem}}} \right),
	\end{cases}
	\label{eq: slmp matrixs} 
\end{equation}
where $ \boldsymbol{W}^{\text{Ins}} \in \mathbb{R}^{N \times M} $ and $ \boldsymbol{P}^{\text{Ins}} \in \mathbb{R}^{M \times N} $ represent the instance-level weight matrix and assignment matrix, respectively. Similarly, $ \boldsymbol{W}^{\text{Sem}} \in \mathbb{R}^{N \times S} $ and $ \boldsymbol{P}^{\text{Sem}} \in \mathbb{R}^{S \times N} $ denote the semantic-level weight matrix and assignment matrix. Based on (\ref{eq: slmp}), we deduce the following equation:
\begin{equation}
	\begin{cases} 
		\boldsymbol{F}^{\text{Pos}} = \boldsymbol{F}^{\text{Ins}} \boldsymbol{W}^{\text{Ins}},
		\\ 
		\boldsymbol{F}^{\text{Geo-i}} = \boldsymbol{F}^{\text{Sem}} \boldsymbol{W}^{\text{Ins}},
		\\ 
	\boldsymbol{F}^{\text{Geo-s}} = \boldsymbol{F}^{\text{Sem}} \boldsymbol{W}^{\text{Sem}}.
	\end{cases}
	\label{eq: slmp features} 
\end{equation}
Here, $\boldsymbol{F}^{\text{Pos}} \in \mathbb{R}^{D \times M}$ represents the pose features of object parts. The geometric features are denoted by $\boldsymbol{F}^{\text{Geo-i}} \in \mathbb{R}^{D \times M}$ for individual parts and $\boldsymbol{F}^{\text{Geo-s}} \in \mathbb{R}^{D \times S}$ for shared geometric features among repeatable parts.

\subsection{Alignment Induced Decoder}
\subsubsection{Feature alignment}
A common understanding is that parts with similar semantics within the same object tend to possess identical or closely related geometric features. In other words, the geometric features of semantically similar parts in $\boldsymbol{F}^{\text{Geo-i}}$ and $\boldsymbol{F}^{\text{Geo-s}}$ should be aligned within the same geometric subspace. A straightforward approach to aligning $\boldsymbol{F}^{\text{Geo-i}}$ and $\boldsymbol{F}^{\text{Geo-s}}$ is to use a shared decoder for predicting primitive parameters through shape reconstruction. However, as discussed in Section \ref{sec: preliminaries dsq}, each primitive in our shape abstraction necessitates the joint definition of both pose and geometric parameters. While instance-level $\boldsymbol{F}^{\text{Geo-i}} \in \mathbb{R}^{D \times M}$ matches pose parameters corresponding to the primitive number $M$, semantic-level $\boldsymbol{F}^{\text{Geo-s}} \in \mathbb{R}^{D \times S}$ lacks this correspondence due to the disparity between $S$ and $M$. Therefore, drawing from the success of attention mechanisms \cite{vaswani2017attention} in various fields for alignment, including image segmentation \cite{zhang2023rethinking}, entity alignment on knowledge graphs \cite{zhang2022benchmark}, and multitasking \cite{zhao2019multiple}, we adopt an attention-based strategy to produce semantic-aligned instance-level geometric features.

\begin{figure}[!tbp]
	\centering
	\includegraphics[width=\linewidth]{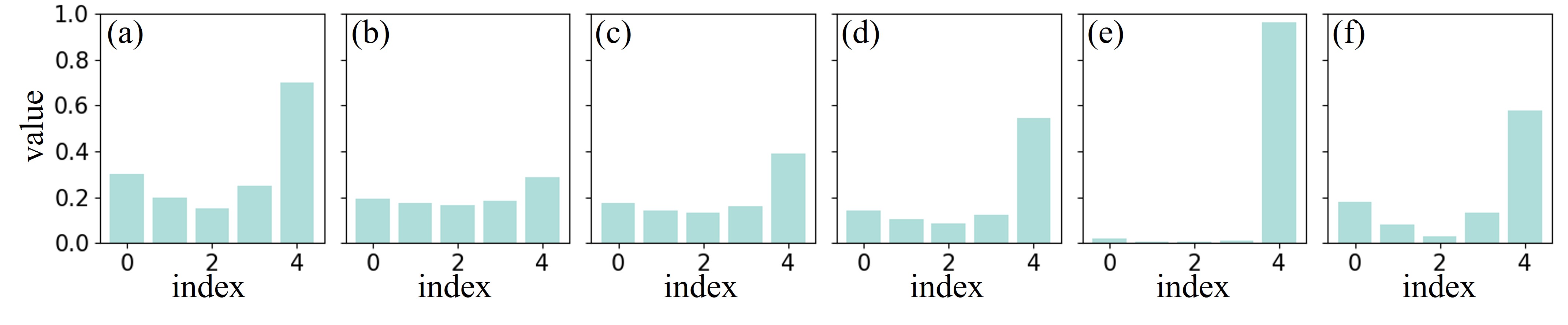}
	\caption{Softmax \emph{vs.} Sparsemax. (a) Input logits. (b) Softmax at temperature 1.0. (c) Softmax at temperature 0.5. (d) Softmax at temperature 0.3. (e) Softmax at temperature 0.1. (f) Sparsemax.}
	\label{fig: functions}
\end{figure}

However, the standard Softmax in the attention tends to disrupt the sparsity of part features by introducing nonzero weights, as illustrated in Fig. \ref{fig: functions} (b). Likewise, Sparsemax in Fig. \ref{fig: functions} (f), devoid of trivial values, faces a similar challenge. Additionally, excessive attention to semantically divergent subspaces undermines the identification of repeatable primitives. To address these issues, we introduce an adaptive temperature $\tau$ into Softmax to provide steep weights:
\begin{equation}
	\boldsymbol{F}^{\text{Geo-is}} \!
	= \!\boldsymbol{F}^{\text{Geo-s}} \!\boldsymbol{W}^{\text{A}}\!
	= \!\boldsymbol{F}^{\text{Geo-s}} \!\operatorname{Softmax}_{\text{col}} \!\left( \!\frac{ \left({\boldsymbol{F}^{\text{Geo-s}}}\right)^{\mathsf{T}}\!\boldsymbol{F}^{\text{Geo-i}}}
	{\tau \sqrt{D}}
	\!\right).
	\label{eq: attention alignment with temperature} 
\end{equation}
Here, $\boldsymbol{F}^{\text{Geo-is}} \in \mathbb{R}^{D \times M}$ represents the semantic-aligned instance-level geometric features, and $\boldsymbol{W}^{\text{A}} \in \mathbb{R}^{S \times M}$ indicates the semantic similarity between individual parts. We determine an adaptive $\tau$ based on the mean square error (MSE) of the cosine distance among semantic-level part geometric features:
\begin{equation}
	\tau = \operatorname{MSE} \left( 
	\left({{\boldsymbol{\hat{F}}}^{\text{Geo-s}}}\right)^{\mathsf{T}} \boldsymbol{\hat{F}}^{\text{Geo-s}},
	\boldsymbol{I}_{S}
	\right),
	\label{eq: tau} 
\end{equation}
where $ \boldsymbol{\hat{F}}^{\text{Geo-s}}$ represents the column-wise $ \ell^2 $-normalized version of $ \boldsymbol{F}^{\text{Geo-s}}$, and $\boldsymbol{I}_{S} $ denotes a $S$-order identity matrix. Our novel $\tau$ ensures maximal orthogonality among semantic subspaces, thus promoting sparsity and alleviating mutual interference.

\subsubsection{Decoder}
\begin{figure}[!tbp]
	\centering
	\includegraphics[width=\linewidth]{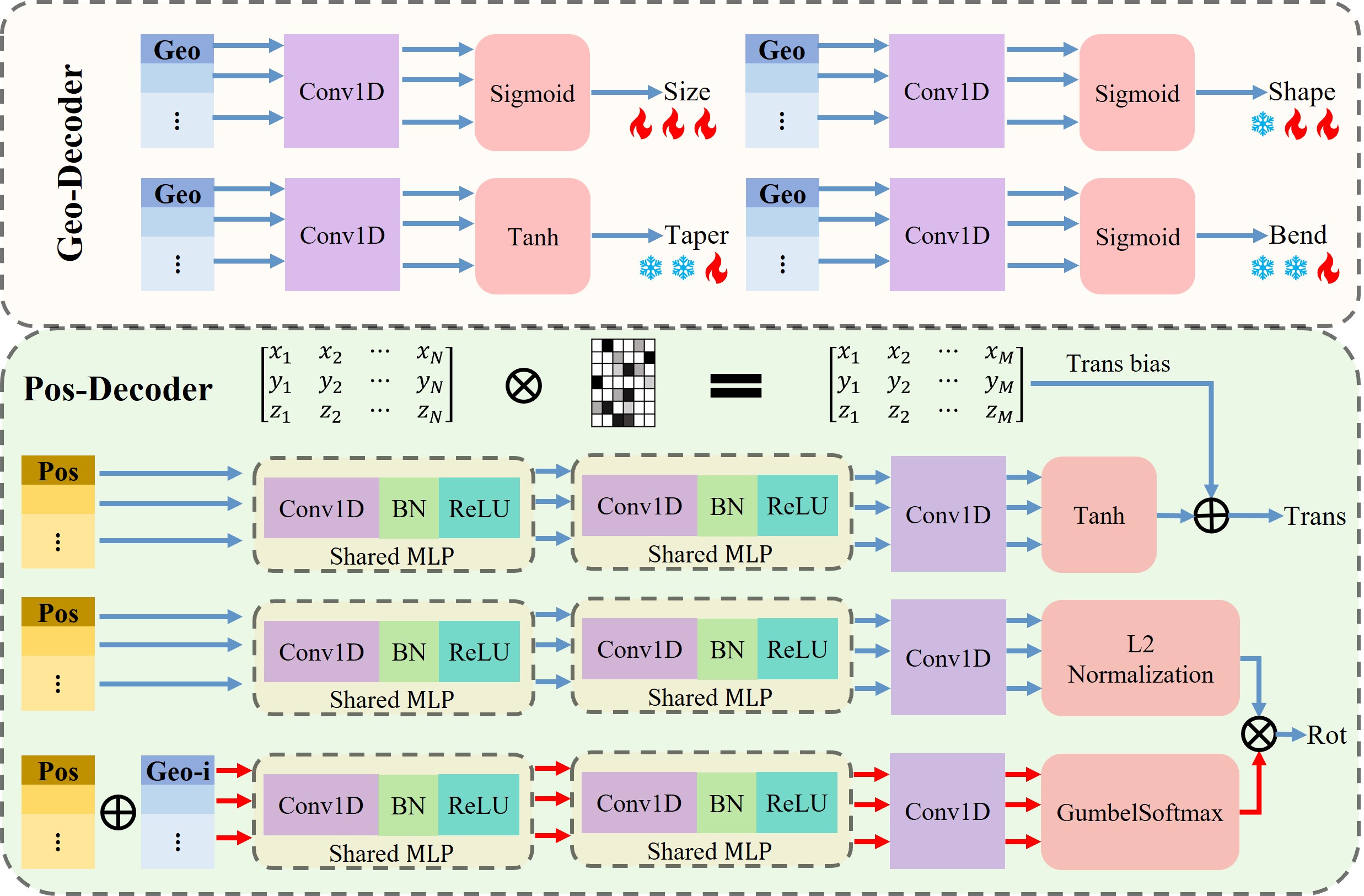}
	\caption{Pos-Decoder and Geo-Decoder for predicting deformable superquadrics. The symbol $\bigotimes$ represents matrix multiplication (for Trans bias) or quaternion multiplication (for Rot), while $\bigoplus$ denotes point-wise addition. Geo refers to any one of Geo-i, Geo-is, or Geo-s, with their respective Geo-Decoders sharing parameters. Flames and snowflakes indicate whether the corresponding parameters are learnable or frozen during the current training process.}
	\label{fig: decoder}
\end{figure}

For the geometric parameters, we employ a single linear mapping achieved through one-dimensional convolution (Conv1D) without any norm layers to predict each parameter type, as shown in Fig. \ref{fig: decoder}. This simplest structure minimizes potential ambiguity in the correspondence between geometric features and parameters. To ensure semantic-geometry consistency among semantically similar primitives, we adopt a cascade unfrozen learning on our primitive (DSQs) parameter decoder (see Fig. \ref{fig: unfreezing} and Fig. \ref{fig: decoder}). Specifically, we start by training only the size parameter, followed by unfreezing the shape parameter, and ultimately unfreezing the bending and tapering parameters. 

Regarding the translation parameter $\boldsymbol{t}$ in the pose parameters, we follow the approach described in LMP \cite{li2024shared}, predicting $\boldsymbol{t}$ using $ \boldsymbol{X}\boldsymbol{W}^{\text{Ins}} $ as a translation bias for robust training. For the rotation parameter $\boldsymbol{r}$, we devise a Gumbel-Softmax-based sub-network (see sub-network with red arrows in Fig. \ref{fig: decoder}) to discern whether parts exhibit mirror relationships and identify their mirror planes. 

\subsubsection{Cascade unfrozen learning}
\begin{figure}[!tbp]
	\centering
	\includegraphics[width=\linewidth]{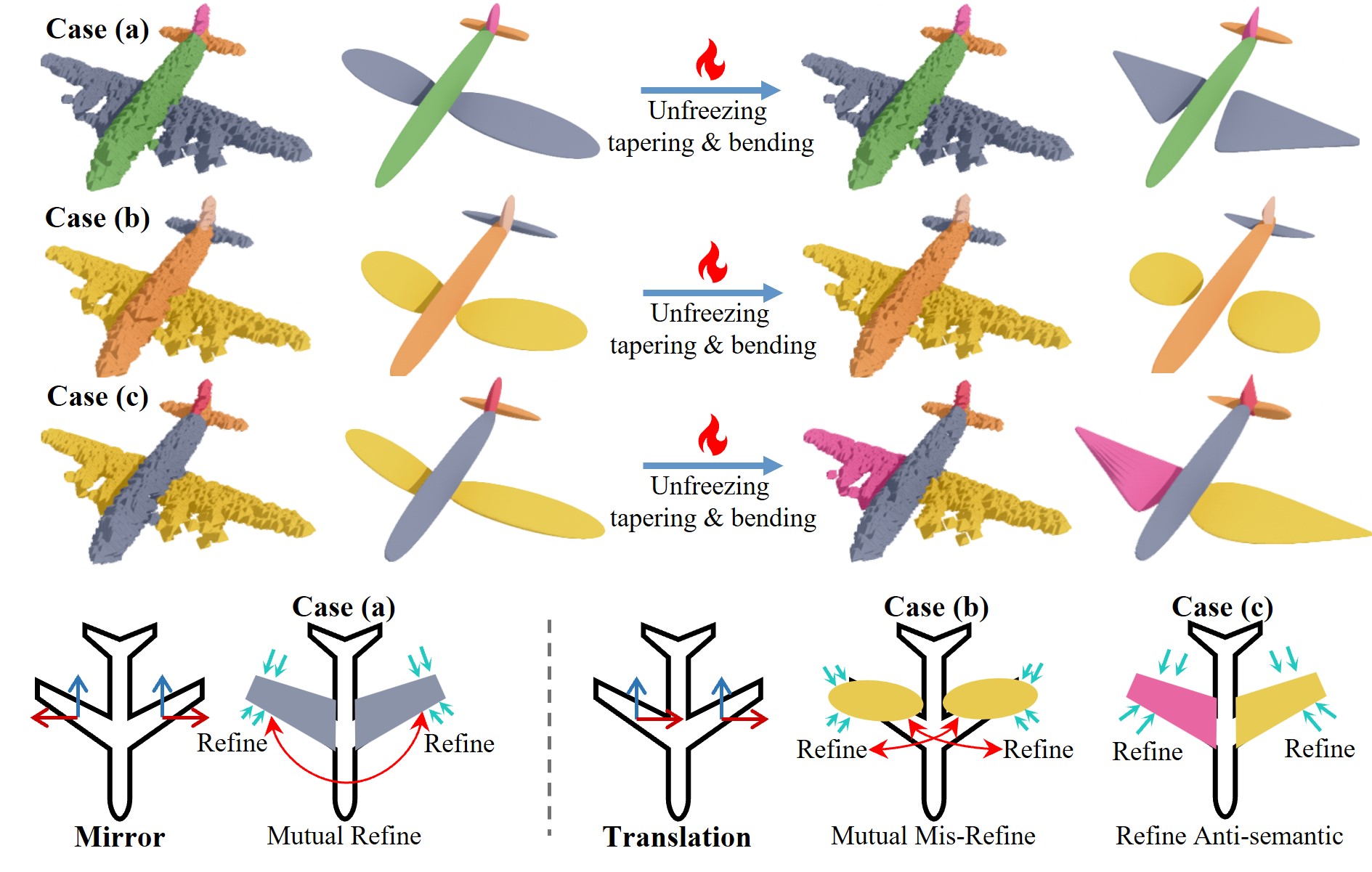}
	\caption{Symmetry identification. The left side of the first three rows presents the results of semantic segmentation and shape abstraction with repeatable primitives (each primitive exhibiting mirror symmetry) when training the size and shape parameters jointly. The right side displays the outcomes when all geometric parameters are trained simultaneously.}
	\label{fig: symmetry}
\end{figure}

When a primitive exhibits mirror symmetry in its geometry, it becomes indifferent to the mirror symmetry and translation symmetry when decoding the pose parameters, which are typically found in man-made object shapes. This results in simplified solutions where mirror symmetry in pose can be represented merely as translation symmetry, with primitives translated along axes without reflection. As shown in Fig. \ref{fig: symmetry}, for airplane wings with frozen tapering and bending parameters, mirror and translation transformations are geometrically indistinguishable. However, upon unfreezing the parameters, correctly identifying the mirror case (a) in Fig. \ref{fig: symmetry} facilitates semantically-supported refinement (tapering and bending). Conversely, misidentifying mirrors as translation in cases (b) and (c) results in two types of degenerate solutions: either maintaining semantic coherence and averaging the geometry for repeatable parts, or breaking semantic for individual refinement. To address this challenge, we employ the Gumbel-Softmax with temperature to predict a 4-dimensional one-hot vector for each primitive, enabling quaternion multiplication to handle mirrors across the $xy$-plane, $xz$-plane, $yz$-plane, and absence of mirrors.

\subsubsection{Outputs}
Based on (\ref{eq: slmp features}) and (\ref{eq: attention alignment with temperature}), we can use $\left\{{\boldsymbol{F}}^{\text{Pos}}, {\boldsymbol{F}}^{\text{Geo-i}} \right\}$ and $\left\{{\boldsymbol{F}}^{\text{Pos}}, {\boldsymbol{F}}^{\text{Geo-is}} \right\}$ to decode instance shape abstraction $\boldsymbol{\Theta}^{\text{Ins}}:=\left\{\boldsymbol{\theta}^{\text{Ins}}_m \in \mathbb{R}^{16} \right\}_{m=1}^M$ and semantic shape abstraction $\boldsymbol{\Theta}^{\text{Sem}}:=\left\{\boldsymbol{\theta}^{\text{Sem}}_m \in \mathbb{R}^{16} \right\}_{m=1}^M$ through shape reconstruction constraints with alignment. For shape abstraction with repeatable primitives, we employ $\left\{{\boldsymbol{F}}^{\text{Pos}}, {\boldsymbol{F}}^{\text{Geo-s}}, \boldsymbol{W}^{\text{A}}\right\}$ to predict $\boldsymbol{\Theta}^{\text{Rep}}:=\left\{\boldsymbol{\theta}^{\text{Rep}}_m \in \mathbb{R}^{16} \right\}_{m=1}^M$. Specifically, semantically unique $S$ primitives are repeated according to the matrix $ \operatorname{Onehot} \left( \operatorname{Argmax}_{\text{col}}\left( \boldsymbol{W}^{\text{A}} \right)\right) \in \mathbb{R}^{S \times M}$, which is composed of one-hot column vectors. Notably, the Argmax is non-differentiable. Instead of directly training repeatable primitives, we derive them naturally via alignment. For segmentation, semantic segmentation are produced from $\operatorname{Argmax}_{\text{col}}\left( \boldsymbol{P}^{\text{Sem}} \right)$, and instance segmentation are derived from $\operatorname{Argmax}_{\text{col}}\left( \boldsymbol{P}^{\text{Ins}} \right)$.

To create adaptive and concise semantic and instance shape abstractions tailored to complex object shapes, we tally the corresponding instance-level segmented points for each primitive. Primitives surpassing a specified threshold (empirically set at 20) in point count are included in the concise shape abstraction, while those with insufficient points are masked. Primitives in semantic shape abstractions share identical existence indicators as those in instance shape abstractions. This non-learning-based statistical approach to implementing concise shape abstractions avoids the necessity for additional loss functions about existence similar to those found in LMP \cite{li2024shared} and CAS \cite{yang2021unsupervised}.

\subsection{Loss Fuction}
We train our unsupervised and one-stage method through shape reconstruction constraints. To efficiently utilize GPU acceleration and measure the reconstruction gap, we utilize PyTorch3D \cite{ravi2020pytorch3d} to uniformly sample $I$ points $\boldsymbol{Y}^m:=\left\{ \boldsymbol{y}^m_i \in \mathbb{R}^{3} \right\}_{i=1}^I $ per primitive $ \boldsymbol{\theta}_m $. 

Our loss function comprises five components, including four pairwise loss functions derived from the losses outlined in LMP \cite{li2024shared}, and an instance-semantic alignment loss based on MSE. To define our losses, we introduce two fundamental distances similar to those in LMP: point-to-primitive and primitive-to-point. The point-to-primitive distance from an input point $\boldsymbol{x}_n$ to a deformable superquadric $\boldsymbol{\theta}_m$ is deﬁned as follows:
\begin{equation}
	d\left(\boldsymbol{x}_n, \boldsymbol{\theta}_m \right) = \min_{i}{\Vert \boldsymbol{x}_n - \boldsymbol{y}_i^m \Vert_2 }.
	\label{eq: pc2dsq} 
\end{equation}
The primitive-to-point distance is defined as the $ \ell^1 $-norm based coverage of a deformable superquadric $ \boldsymbol{\theta}_m$ over its covering points:
\begin{equation}
	c \left( \boldsymbol{\theta}_m \right) = \frac{1}{I}\sum_{i=1}^{I}{\min_{n}{\Vert \boldsymbol{y}_i^m - \boldsymbol{x}_n \Vert_2 }}.
	\label{eq: dsq2pc} 
\end{equation}
Based on the definitions above, we derive the following losses.

\subsubsection{Anti-anchor loss}
\label{sec: ls_hd}
We utilize the bidirectional Hausdorff distance to expedite training and encourage the model to explore various assembly schemes :
\begin{equation}
	\operatorname{HD}\!\left(\!\Theta, \!\boldsymbol{X}\!\right)\! = \!
	\max\!\big( \!
	\max_{n}\min_{m}d\left(\!\boldsymbol{x}_n, \!\boldsymbol{\theta}_m\!\right),
	\max_{m}\min_{n}d\left(\!\boldsymbol{x}_n, \!\boldsymbol{\theta}_m\!\right)
	\!\big),
	\label{eq: hd}
\end{equation}
\begin{equation}
	\mathcal{L}_{hd} = 0.5 * \operatorname{HD}\left(\Theta^\text{Ins}, \boldsymbol{X} \right) + 
	0.5 * \operatorname{HD}\left(\Theta^\text{Sem}, \boldsymbol{X} \right).
	\label{eq: ls_hd} 
\end{equation}

\subsubsection{Reconstruction loss}
\label{sec: ls_recon}
Our shape reconstruction loss is written as
\begin{equation}
	\operatorname{Recon} \!\left(\!\Theta, \!\boldsymbol{X}\!\right) \!= \!\frac{1}{2N}\! \sum_{n=1}^{N}{\min_{m}{ d\left(\!\boldsymbol{x}_n, \!\boldsymbol{\theta}_m \!\right)}}\! +\! \frac{1}{2M}\!\sum_{m=1}^{M}{ c \left( \boldsymbol{\theta}_m \right) },
	\label{eq: recon}
\end{equation}
\begin{equation}
	\mathcal{L}_{\text{recon}} = 
	0.5 * \operatorname{Recon}\left(\!\Theta^\text{Ins}, \!\boldsymbol{X} \!\right) 
	\!+ \!
	0.5 * \operatorname{Recon}\left(\!\Theta^\text{Sem}, \!\boldsymbol{X} \!\right).
	\label{eq: ls_recon}
\end{equation}

\subsubsection{Anti-collapse loss}
\label{sec: ls_wd}
To prevent misassignment in segmentation caused by overlapping primitives, we approach point cloud segmentation as an optimal transportation problem. We introduce an anti-collapse loss to ensure precise point-to-primitive membership and prevent segmenting points into trivial single-semantic segments. The distinction from previous approaches \cite{yang2021unsupervised, li2024shared} lies in our incorporation of a regularization term $\delta_\text{wd}=0.05$ to mitigate the influence of minor perturbations in the input point cloud, thereby preventing over-segmentation. Our anti-collapse loss is formulated as
\begin{equation}
	\hat{d}\left(\cdot\right) = \mathbb{I} \left( d\left(\cdot\right) > \delta_\text{wd}\right) d\left(\cdot\right),
	\label{eq: regularization}
\end{equation}
\begin{equation}
		\mathcal{L}_{\text{wd}} = \frac{1}{N}\sum_{n=1}^{N}{\sum_{m=1}^{M} p^{\text{Ins}}_{mn} \left(
			\frac{
				\hat{d}\left(\boldsymbol{x}_n, \boldsymbol{\theta}^\text{Ins}_m \right) 
				+
				\hat{d}\left(\boldsymbol{x}_n, \boldsymbol{\theta}^\text{Sem}_m \right)
			}
			{2}
			\right) },
	\label{eq: ls_wd} 
\end{equation}
where $p^{\text{Ins}}_{mn}$ denotes an element of the instance-level assignment matrix $\boldsymbol{P}^{\text{Ins}}$ as defined in (\ref{eq: slmp matrixs}), and $\mathbb{I} (\cdot) $ is the indicator function.

\subsubsection{Compactness loss}
\label{sec: ls_compact}
To produce compact shape abstractions, similar to the approach in \cite{yang2021unsupervised, li2024shared}, we incorporate a regularization term where $\delta_\text{c}=0.01$:
\begin{equation}
	\mathcal{L}_{\text{c}}= \left(\frac{1}{M} \sum_{m=1}^{M}{\sqrt{\frac{1}{N}\sum_{n=1}^{N}{p^{\text{Ins}}_{mn}} + \delta_\text{c}}} \right)^2.
	\label{eq: ls_sp} 
\end{equation}

\subsubsection{Alignment loss}
\label{sec: ls_alignment}
We derive both semantic-level and instance-level memberships by disentangling pose features and geometric features. While instance-level membership is reliably captured through constraints $\mathcal{L}_{\text{wd}}$ and $\mathcal{L}_{\text{c}}$, semantic-level membership struggles to establish robust relationships. To rectify erroneous relationships in semantic-level membership, we generate pseudo-labels from the similarity matrix $\boldsymbol{W}^\text{A}$ obtained in (\ref{eq: attention alignment with temperature}) to align $\boldsymbol{P}^{\text{Ins}}$ and $\boldsymbol{P}^{\text{Sem}}$. This alignment allows instance-level membership to effectively inform and refine semantic-level membership. The formulation is as follows:
\begin{equation}
	\boldsymbol{P}^{\text{Sem-pseudo}} = \boldsymbol{\hat{W}}^{\text{A}} \boldsymbol{P}^{\text{Ins}}, 
	\hat{W}^{\text{A}}_{sm}= 
	\begin{cases}
	1, & \text{if $W^{\text{A}}_{sm} = \max_{s}{W^{\text{A}}_{sm}}$},\\ 
	0, &{\text{otherwise}},
	\end{cases}
	\label{eq: align pseud} 
\end{equation}
\begin{equation}
	\mathcal{L}_{\text{a}}= \operatorname{MSE}\left(
	\boldsymbol{P}^{\text{Sem}}, \boldsymbol{P}^{\text{Sem-pseudo}}
	\right).
	\label{eq: ls_align} 
\end{equation}

The overall loss is a weighted sum of these loss functions:
\begin{equation}
	\mathcal{L}= \mathcal{L}_{\text{recon}} + \lambda_1 \mathcal{L}_{\text{hd}} + \lambda_2 \mathcal{L}_{\text{wd}} + \lambda_3 \mathcal{L}_{\text{c}} + \lambda_4 \mathcal{L}_{\text{a}}.
	\label{eq: ls_all} 
\end{equation}

\section{Experiments}
\label{sec: experiments}
\subsubsection{Implementation details}
For all shape categories, we set the maximum primitive number as $M=16$, and the maximum semantic number as $S=6$. We uniformly sample $I = 256$ points on each primitive mesh, following \cite{li2024shared}. Our encoder leverages the segmentation architecture of PointNet++ \cite{qi2017pointnet++} to map a point cloud into a 384-dimensional latent space, resulting in $D=192$ dimensions for both pose features and geometric features. Additionally, to ensure a stable mapping between features and geometric parameters, avoiding semantic ambiguity and ensuring numerical stability for gradient propagation, we impose constraints on the parameter space of deformable superquadrics, as detailed in TABLE \ref{tab: parameters}.

\subsubsection{Dataset}
Building on previous approaches \cite{tulsiani2017learning, sun2019learning, paschalidou2019superquadrics, yang2021unsupervised, li2024shared}, we conduct experiments on four shape categories: airplane (3640 instances), chair (5929), and table (7555) from ShapeNet \cite{chang2015shapenet}, and four-legged animal (129) from Animal \cite{tulsiani2017learning}. Each point cloud is normalized to unit scale and aligned within a canonical frame. The training and testing sets are divided in a $4:1$ ratio. Our model does not require separate training for each category.

\subsubsection{Network training}
All experiments are conducted over 500 epochs on an NVIDIA Geforce GTX3090 GPU using the AdamW optimizer \cite{loshchilov2019decoupled}. We use a batch size of 112, an initial learning rate of $1e-3$ with a cosine decay scheduler to reduce to $3e-4$ by the end of training, and a weight decay of $1e-3$. The weights of our loss function are set as follows: $\lambda_1 = 1$, $\lambda_2 = 0.3$, $\lambda_3 = 0.1$, and $\lambda_4 = 0.01$, using the sum-over-batch approach. During training, we employ the cascade unfreezing strategy to our Geo-Decoder. Specifically, the initial 100 epochs focuse exclusively on training with cuboid-like superquadrics, followed by 100 epochs dedicated to non-deformable superquadrics, and concluding with 300 epochs on deformable superquadrics. Over the initial 100 epochs, $\lambda_2$ doubles and $\lambda_3$ triples, after which they revert to their original values. To ensure semantic-geometry consistency, while mitigating potential disruption caused by excessive local geometry refinement through the Hausdorff distance metric, we reduce $\lambda_1$ to 0 after the 50th epoch and maintain this adjustment throughout the training process. The network implementation utilized the PyTorch framework.

\begin{table*}
	\renewcommand\arraystretch{1}
	\begin{center}
		\caption{Summary of our method compared to baselines. Methods marked with an asterisk (*) indicate unofficial implementations. SQEMS and SQBI are non-learning-based methods, while the others are learning-based methods.}
		\label{tab: baselines}
		\begin{tabular}{| c | c | c | c | c | c | c | c | c | c | }
			\hline
			Method & Publication & Year & Category Free & One Stage & Adaptive Parts & Semantics & Repeatable Parts & Primitive & Output \\ \hline
			CA \cite{sun2019learning}
			& TOG & 2019 &\usym{2717} & \usym{2717} & \checkmark & \usym{2717} & \usym{2717} & cuboid & mesh \\ \hline
			SQA \cite{paschalidou2019superquadrics} 
			& CVPR & 2019 & \usym{2717}& \checkmark & \checkmark & \usym{2717} & \usym{2717} & SQ & mesh \\ \hline
			CAS \cite{yang2021unsupervised} 
			& TOG & 2021 &\usym{2717} & \checkmark & \checkmark & \usym{2717} & \usym{2717} & cuboid & mesh \& points\\ \hline
			SQEMS \cite{liu2022robust} 
			& CVPR & 2022 & - & - & \checkmark &\usym{2717} & \usym{2717} & SQ & mesh \& points \\ \hline
			SQBI \cite{wu2022primitive} 
			& ECCV & 2022 & - & - & \checkmark & \usym{2717} & \usym{2717} & DSQ & mesh \& points \\ \hline
			EditVAE \cite{li2022editvae} 
			& AAAI & 2022 &\usym{2717}& \checkmark & \usym{2717} & \checkmark & \usym{2717} & DSQ & points \\ \hline
			ProGRIP* \cite{deng2022unsupervised} 
			& NeurIPS & 2022 &\usym{2717}& \usym{2717} & \checkmark & \checkmark & \checkmark & cuboid & mesh \\ \hline
			LMP \cite{li2024shared} 
			& TIP & 2024 &\checkmark& \checkmark & \checkmark & \usym{2717} & \usym{2717} & DSQ & mesh \& points \\ \hline
			CoPartSeg* \cite{umam2024unsupervised} 
			& TMM & 2024 &\usym{2717}& \usym{2717} & \usym{2717} & \checkmark & \usym{2717} & DSQ & points \\ \hline
			Ours
			& - & - &\checkmark& \checkmark & \checkmark & \checkmark & \checkmark & DSQ & mesh \& points \\ \hline
		\end{tabular}
	\end{center}
\end{table*}

\subsubsection{Baselines}
\label{sec: reconstruction baslines}
We demonstrate the performance of our fully unsupervised, one-stage, learning-based method for semantic representation of man-made 3D shapes, through both qualitative and quantitative comparisons with 9 state-of-the-art (SOTA) primitive-based shape representation approaches: CA \cite{sun2019learning}, SQA \cite{paschalidou2019superquadrics}, CAS \cite{yang2021unsupervised}, SQEMS \cite{liu2022robust}, SQBI \cite{wu2022primitive}, EditVAE \cite{li2022editvae}, ProGRIP \cite{deng2022unsupervised}, LMP \cite{li2024shared}, and CoPartSeg \cite{umam2024unsupervised}. A summary of these comparisons is presented in TABLE \ref{tab: baselines}. CA employs an auto-encoder (AE)-based cuboid prediction network in its first stage to generate hierarchical cuboid abstractions. Then, the network is fixed to train the cuboid selection module. CA requires point normals for weakly-supervised shape reconstruction, and input point clouds must belong to the same category. SQA utilizes DSQs as primitives and introduces the Parsimony Loss, eliminating the necessity for the selection module used in CA. Similarly, SQA also requires category-specific oriented point clouds for training. With similar requirements, CAS combine instance-level shape abstraction and segmentation using a probabilistic point-to-cuboid allocation matrix through the attention mechanism. SQEMS segments the point cloud before approximating grouped local points with SQs, whereas SQBI generates a DSQs-based shape abstraction before conducting clustering-based segmentation. EditVAE predefines a small maximum numbers of primitives (\emph{e.g.}, 3) per category to produce semantic, limiting flexibility and interpretability. Utilizing cuboids predicted by the first-staged trained CAS as the target shape, ProGRIP predicts geometric parameters for 10 independent cuboids and 6 pose parameters for each cuboid, resulting in a total of 60 cuboids for reconstructing the target to discover repeatable parts. Subsequently, ProGRIP refines cuboid-based shapes using GT occupancy functions, making it fundamentally a three-stage method. We select ProGRIP second-stage outputs for comparison in the shape abstraction task. LMP unites instance-level shape abstraction and segmentation through the shared instance-level membership, eliminating the need for separate models per point cloud category. In the first stage, CoPartSeg \footnote{Since the official code is not yet publicly available, we adjusted the loss function weights to aid convergence and added cross-entropy loss in the second stage to stabilize training for the airplane, chair, and table categories.} reconstructs superpoints from 15 DSQs. Then, it performs mean pooling using the superpoint labels to derive new superpoint features, which are combined with learnable part tokens (3 for airplanes and chairs, and 2 for tables) for semantic segmentation.

\subsection{Unsupervised Shape Representation}
\subsubsection{Semantic segmentation}
\label{sec: semantic segmentation}
We evaluate the semantic segmentation performance against EditVAE \cite{li2022editvae}, ProGRIP \cite{deng2022unsupervised} and CoPartSeg \cite{umam2024unsupervised}. Specifically, the second stage of ProGRIP does not produce semantic segmentation. As the stage relies on instance-level cuboid-based shape abstraction from CAS \cite{yang2021unsupervised} as GT for shape reconstruction, we establish a one-to-one correspondence between the cuboids predicted by ProGRIP and those from CAS. Consequently, we transfer the instance segmentation from CAS to the semantic segmentation of ProGRIP. Meanwhile, the ground truth (GT) semantic labels for airplanes, chairs and tables are sourced from ShapeNetCore \cite{yi2016scalable}, while animal point clouds lack associated GT labels. To ensure a comprehensive quantitative comparison of point clouds with and without GT labels, we combine the commonly used supervised segmentation performance metric mean Intersection over Union (mIoU) \cite{guo2021deep} with clustering metrics, such as Normalized Mutual Information (NMI) and Davies-Bouldin Index (DBI) \cite{fahad2014survey}. mIoU gauges the diversity and similarity between segmented point groups and GT point groups, where higher values signify improved performance. NMI, an external clustering metric that requires GT, quantifies mutual information between predictions and GT to determine their similarity, with higher values indicating superior similarity and shared information. DBI, an internal clustering metric independent of GT, evaluates discrepancy between each segmented point group and its most similar counterpart.

\begin{figure*}[!tp]
	\centering
	\includegraphics[width=\textwidth]{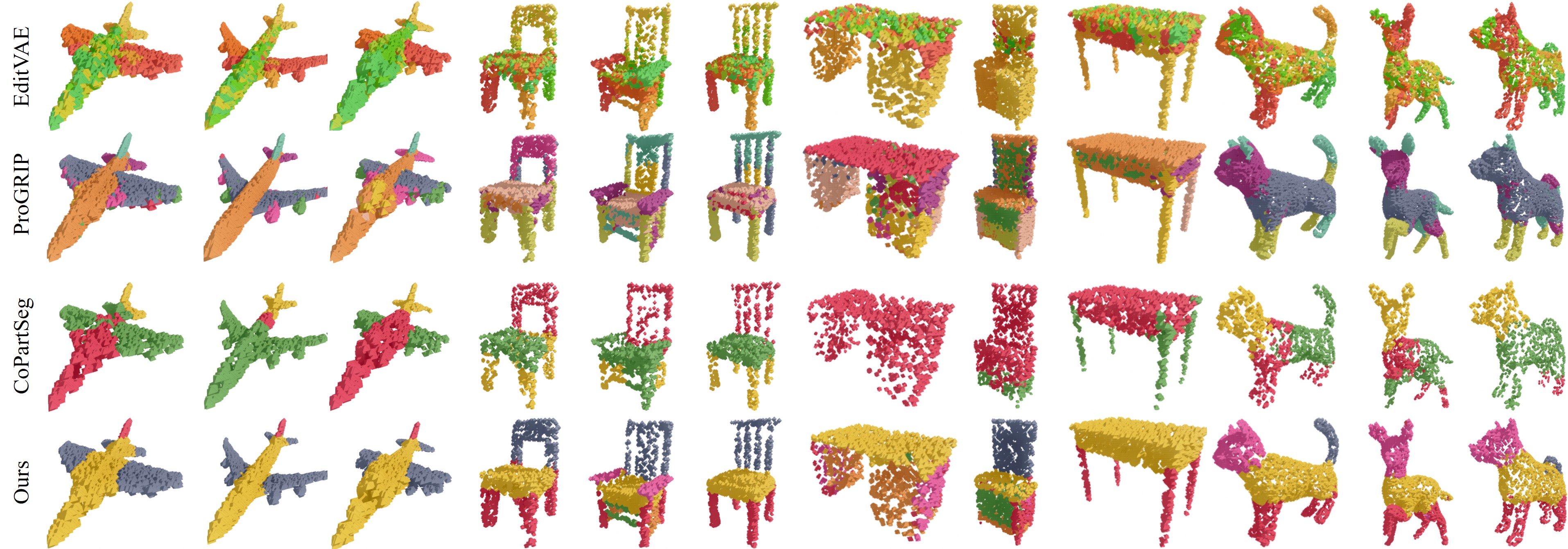}
	\caption{
		Comparison of semantic segmentation results. ProGRIP \cite{deng2022unsupervised} uses 4096 points with normal information in stage one and 4096 points in stage two. CoPartSeg \cite{umam2024unsupervised} uniformly samples 1024 points. Other methods input 2048 points each. Our approach identifies and highlights the main components of objects across different categories using yellow coding.
	}
	\label{fig: semantic segmentation}
\end{figure*}

\begin{table}
	\renewcommand\arraystretch{1.2}
	\begin{center}
		\caption{Semantic Segmentation Quantitative Comparison: mean Intersection over Union (mIoU), Normalized Mutual Information (NMI) and Davies-Bouldin Index (DBI).}
		\label{tab: semantic segmentation}
		\begin{tabular}{|c|cp{0.1cm}|c|c|c|c|}
			\hline
			Metric & \multicolumn{2}{c|}{Method} & Airplane & Chair & Table & Animal \\ \hline
			\multirow{4}{*}{mIoU$\uparrow$}
			&\multicolumn{2}{c|}{EditVAE \cite{li2022editvae}}
			& 0.1869 & 0.1176 & 0.1552 & - \\ \cline{2-7}
			&\multicolumn{2}{c|}{ProGRIP \cite{deng2022unsupervised}}
			& 0.5120 & 0.1228 & 0.2326 & - \\ \cline{2-7}
			&\multicolumn{2}{c|}{CoPartSeg \cite{umam2024unsupervised}}
			& 0.5494 & 0.1801 & \pmb{0.4660} & - \\ \cline{2-7}
			&\multicolumn{2}{c|}{Ours}
			& \pmb{0.6134} & \pmb{0.2069} & 0.4405 & -\\ \cline{2-7}
			\hline
			\multirow{4}{*}{NMI$\uparrow$}
			&\multicolumn{2}{c|}{EditVAE \cite{li2022editvae}}
			& 0.4056 & 0.4081 & 0.2581 & - \\ \cline{2-7}
			&\multicolumn{2}{c|}{ProGRIP \cite{deng2022unsupervised}}
			& \pmb{0.6259} & 0.5582 & 0.4662 & - \\ \cline{2-7}
			&\multicolumn{2}{c|}{CoPartSeg \cite{umam2024unsupervised}}
			& 0.4051 & 0.4317 & 0.2945 & - \\ \cline{2-7}
			&\multicolumn{2}{c|}{Ours}
			& 0.5560 & \pmb{0.6496} &\pmb{0.5272} & - \\ \cline{2-7}
			\hline
			\multirow{4}{*}{DBI$\downarrow$}
			&\multicolumn{2}{c|}{EditVAE \cite{li2022editvae}}
			& 8.577 & 5.448 & 20.39 & 6.652 \\ \cline{2-7}
			&\multicolumn{2}{c|}{ProGRIP \cite{deng2022unsupervised}}
			& 5.251 & 3.951 & 4.271 & 3.331 \\ \cline{2-7}
			&\multicolumn{2}{c|}{CoPartSeg \cite{umam2024unsupervised}}
			& \pmb{4.237} &2.643 & \pmb{2.267} & 2.531 \\ \cline{2-7}
			&\multicolumn{2}{c|}{Ours}
			& 4.820 & \pmb{2.584} & 3.812 & \pmb{1.345} \\ \cline{2-7}
			\hline
		\end{tabular}
	\end{center}
\end{table}

TABLE \ref{tab: semantic segmentation} presents the quantitative results of semantic segmentation, while Fig. \ref{fig: semantic segmentation} illustrates the qualitative comparisons. In metrics using GT labels, such as mIoU and NMI, our method consistently ranks first and occasionally second. Notably, the first stage of ProGRIP, CAS, requires point normals during training. The additional input gives ProGRIP and CAS an advantage in identifying small-scale parts due to sharp normal changes at the junctions between parts of man-made objects, as seen with animal ears and airplane engines in Fig. \ref{fig: semantic segmentation}. CoPartSeg, however, predefines a minimal number of semantics, segmenting airplanes and chairs into three and tables into two. This critical and restrictive predefined hyperparameter allows CoPartSeg to excel in the DBI but results in spurious cross-category identification of object main bodies. In contrast, our method, with a sufficient number of semantics, autonomously identifies the main bodies of objects in cross-category training.

\subsubsection{Instance segmentation}
We compare the instance segmentation performance with CAS \cite{yang2021unsupervised}, SQEMS \cite{liu2022robust}, SQBI \cite{wu2022primitive} and LMP \cite{li2024shared}. As in Section \ref{sec: semantic segmentation}, we quantitatively assess instance segmentation using mIoU, NMI and DBI. Unlike the semantic GT labels, ShapeNet \cite{chang2015shapenet} and PartNet \cite{mo2019partnet} do not provide instance GT labels for airplanes. Therefore, we employ iterative binary clustering (k-means) on the semantic-level ShapeNetCore \cite{yi2016scalable} to produce instance labels. In contrast, the labels for chairs and tables are directly sourced from the most detailed instance-level (level-3) of PartNet. Animal point clouds lack GT labels.

\begin{table}
	\renewcommand\arraystretch{1.2}
	\begin{center}
		\caption{Instance Segmentation Quantitative Comparison: mean Intersection over Union (mIoU), Normalized Mutual Information (NMI) and Davies-Bouldin Index (DBI).}
		\label{tab: instance segmentation}
		\begin{tabular}{|c|cp{0.1cm}|c|c|c|c|}
			\hline
			Metric & \multicolumn{2}{c|}{Method} & Airplane & Chair & Table & Animal \\ \hline
			\multirow{5}{*}{mIoU$\uparrow$}
			& \multicolumn{2}{c|}{CAS \cite{yang2021unsupervised}} 
			& 0.5189 & 0.5169 & 0.4462 & -\\ \cline{2-7}
			&\multicolumn{2}{c|}{SQEMS \cite{liu2022robust}} 
			& 0.3560 & 0.3219 & 0.2604 & -\\ \cline{2-7}
			& \multicolumn{2}{c|}{SQBI \cite{wu2022primitive}}
			& 0.3070 & 0.4220 & 0.3817 & -\\ \cline{2-7}
			&\multicolumn{2}{c|}{LMP \cite{li2024shared} }
			& \pmb{0.6565} & 0.5799 &0.4982 & - \\ \cline{2-7}
			&\multicolumn{2}{c|}{Ours}
			& 0.5756 & \pmb{0.6386} & \pmb{0.5399} & -\\ \cline{2-7}
			\hline
			\multirow{5}{*}{NMI$\uparrow$}
			& \multicolumn{2}{c|}{CAS \cite{yang2021unsupervised}} 
			& \pmb{0.6816} & 0.6673 & \pmb{0.6287} & - \\ \cline{2-7}
			&\multicolumn{2}{c|}{SQEMS \cite{liu2022robust}} 
			& 0.4059 & 0.3676 & 0.2742 & - \\ \cline{2-7}
			& \multicolumn{2}{c|}{SQBI \cite{wu2022primitive}}
			& 0.5516 & 0.5902 & 0.5252 & - \\ \cline{2-7}
			&\multicolumn{2}{c|}{LMP \cite{li2024shared} }
			& 0.6719 & \pmb{0.6676} &0.6131 & - \\ \cline{2-7}
			&\multicolumn{2}{c|}{Ours}
			& 0.5918 & 0.6638 & 0.5977 & -\\ \cline{2-7}
			\hline
			\multirow{5}{*}{DBI$\downarrow$}
			& \multicolumn{2}{c|}{CAS \cite{yang2021unsupervised}} 
			& 1.837 & 1.304 & 1.793 & 1.160\\ \cline{2-7}
			&\multicolumn{2}{c|}{SQEMS \cite{liu2022robust}} 
			& 3.317 & 2.198 & 2.087 & 1.858 \\ \cline{2-7}
			& \multicolumn{2}{c|}{SQBI \cite{wu2022primitive}}
			& 1.889 & 1.602 & 1.926 & 1.309 \\ \cline{2-7}
			&\multicolumn{2}{c|}{LMP \cite{li2024shared} }
			& 1.278 & 1.043 &1.227 & 0.997 \\ \cline{2-7}
			&\multicolumn{2}{c|}{Ours}
			& \pmb{1.106} & \pmb{1.005} & \pmb{1.220} & \pmb{0.959}\\ \cline{2-7}
			\hline
		\end{tabular}
	\end{center}
\end{table}

\begin{figure*}[!tp]
	\centering
	\includegraphics[width=\textwidth]{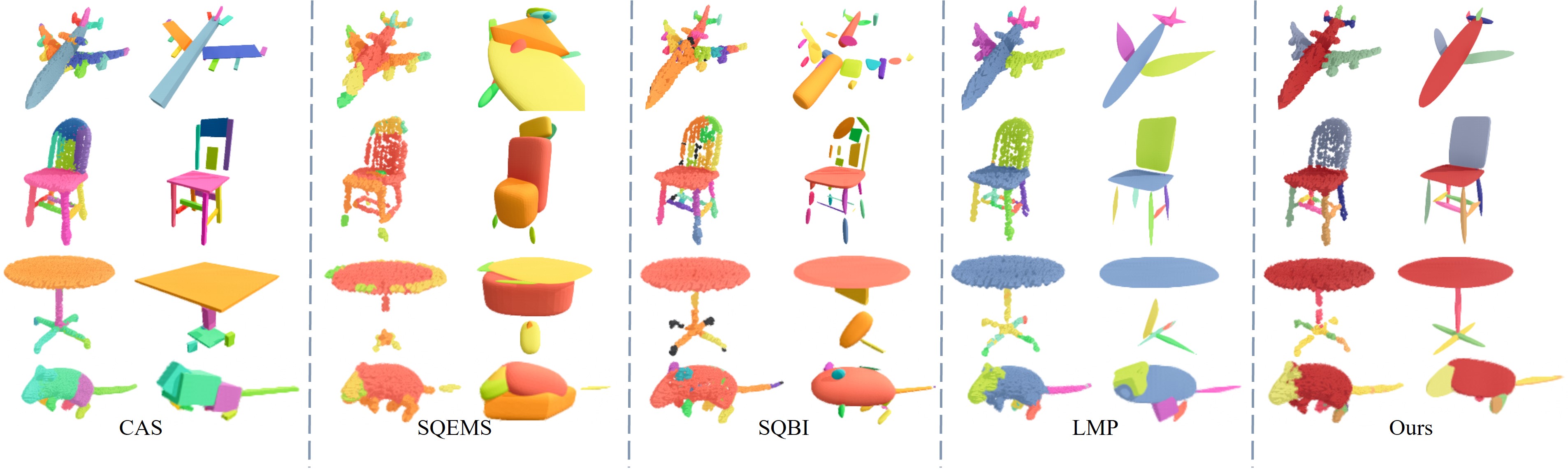}
	\caption{
		Comparison of instance segmentation results. CAS \cite{yang2021unsupervised} uses 4096 points with normal information to identify small-scale parts. As unlearning methods, SQEMS \cite{liu2022robust} segments the downsampled point cloud before produces SQs-based parts, while SQBI \cite{wu2022primitive} generates a DSQs-based shape abstraction prior to segmentation. LMP \cite{li2024shared} unites instance-level shape abstraction and segmentation through the shared instance-level membership for mutual improvement. Both our approach and LMP are trained across various categories.
	}
	\label{fig: instance segmentation}
\end{figure*}

TABLE \ref{tab: instance segmentation} displays the quantitative results of instance segmentation, whereas Fig. \ref{fig: instance segmentation} depicts the qualitative comparisons. CAS, which leverages normal information, excels at segmenting small-scale object parts, such as airplane engines (Fig. \ref{fig: instance segmentation}). The non-learnable SQEMS and SQBI struggle to maintain consistent part interpretations across objects within the same category. While LMP ensures consistent instance-level interpretations across categories, it fails to exploit shared geometric features among semantically similar parts, limiting its effectiveness on objects like chairs and tables (Fig. \ref{fig: instance segmentation}). Quantitatively, our method achieves the highest mIoU and DBI across most categories while remaining highly competitive in NMI. Notably, instance segmentation prioritizes distinguishing individual parts, whereas our approach focuses on capturing semantic information across parts. Despite this difference, both qualitative and quantitative results demonstrate the effectiveness of our method.

\subsubsection{Shape abstraction}
We showcase the performance of our instance shape abstraction, semantic shape abstraction and shape abstraction with repeatable DSQs through both qualitative and quantitative comparisons with 7 state-of-the-art (SOTA) shape abstraction methods: CA \cite{sun2019learning}, SQA \cite{paschalidou2019superquadrics}, CAS \cite{yang2021unsupervised}, SQEMS \cite{liu2022robust}, SQBI \cite{wu2022primitive}, LMP \cite{li2024shared} and ProGRIP \cite{deng2022unsupervised}. To evaluate shape abstraction, we employ the widely adopted $ \ell^2 $-norm Chamfer Distance (CD) and Earth Mover’s Distance (EMD) \cite{sun2019learning, yang2021unsupervised, paschalidou2019superquadrics, li2024shared}, which measure the gap between input and output points. The output points are uniformly sampled from the corresponding derived compact shape abstraction.

\begin{table}
	\renewcommand\arraystretch{1.2}
	\begin{center}
		\caption{Shape reconstruction quantitative comparison: Chamfer Distance (CD-$ \ell^2 $) and Earth Mover’s Distance (EMD). Ours-I denotes instance shape abstraction, Ours-R refers to shape abstraction with repeatable DSQs, and Ours-S signifies semantic shape abstraction.}
		\label{tab: shape abstraction}
		\begin{tabular}{|c|cp{0.1cm}|c|c|c|c|}
			\hline
			Metric & \multicolumn{2}{c|}{Method} & Airplane & Chair & Table & Animal \\ \hline
			\multirow{10}{*}{CD$\downarrow$}
			& \multicolumn{2}{c|}{CA \cite{sun2019learning}}
			& 0.0124 & 0.0050 & 0.0060 & 0.0201 \\ \cline{2-7} 
			& \multicolumn{2}{c|}{SQA \cite{paschalidou2019superquadrics}}
			& 0.0043 & 0.0034 & 0.0036 & 0.0168 \\ \cline{2-7} 
			& \multicolumn{2}{c|}{CAS \cite{yang2021unsupervised}}
			& 0.0013 & 0.0018 & 0.0030 & 0.0014 \\ \cline{2-7} 
			& \multicolumn{2}{c|}{SQEMS \cite{liu2022robust}}
			& 0.0022 & 0.0072 & 0.0054 & 0.0021 \\ \cline{2-7} 
			& \multicolumn{2}{c|}{SQBI \cite{wu2022primitive}}
			& 0.0029 & 0.0027 & 0.0044 & 0.0059 \\ \cline{2-7} 
			& \multicolumn{2}{c|}{LMP \cite{li2024shared}}
			& \pmb{0.0006} & \pmb{0.0008} &\pmb{0.0010} & \pmb{0.0010} \\ \cline{2-7}
			& \multicolumn{2}{c|}{ ProGRIP \cite{deng2022unsupervised}}
			& 0.0034 & 0.0073 & 0.0085 & 0.0109 \\ \cline{2-7} 
			&\multicolumn{2}{c|}{Ours-I}
			& \pmb{0.0006} & 0.0009 & \pmb{0.0010} & 0.0011 \\ \cline{2-7}
			&\multicolumn{2}{c|}{Ours-R}
			& \pmb{0.0006} & 0.0009 & 0.0011 & 0.0011 \\ \cline{2-7}
			&\multicolumn{2}{c|}{Ours-S}
			& \pmb{0.0006} & 0.0009 & \pmb{0.0010} & 0.0011\\ \cline{2-7}
			\hline
			\multirow{10}{*}{EMD$\downarrow$}
			& \multicolumn{2}{c|}{CA \cite{sun2019learning}}
			& 0.0818 & 0.0383 & 0.0366 & 0.0893 \\\cline{2-7} 
			& \multicolumn{2}{c|}{SQA \cite{paschalidou2019superquadrics}}
			& 0.0538 & 0.0299 & 0.0293 & 0.0752 \\ \cline{2-7} 
			& \multicolumn{2}{c|}{CAS \cite{yang2021unsupervised}}
			& 0.0133 &0.0183 & 0.0223 & 0.0168 \\ \cline{2-7} 
			& \multicolumn{2}{c|}{SQEMS \cite{liu2022robust}}
			& 0.0173 & 0.0332 & 0.0294 & 0.0264 \\ \cline{2-7} 
			& \multicolumn{2}{c|}{SQBI \cite{wu2022primitive}}
			& 0.0237 & 0.0204 & 0.0272 & 0.0324 \\ \cline{2-7} 
			& \multicolumn{2}{c|}{LMP \cite{li2024shared}}
			& \pmb{0.0122} & 0.0188 & 0.0210 & \pmb{0.0147} \\ \cline{2-7}
			& \multicolumn{2}{c|}{ ProGRIP \cite{deng2022unsupervised}}
			& 0.0255 & 0.0641 & 0.0360 & 0.0322 \\ \cline{2-7} 
			&\multicolumn{2}{c|}{Ours-I}
			& 0.0130 & \pmb{0.0156} & 0.0202 & 0.0156 \\ \cline{2-7}
			&\multicolumn{2}{c|}{Ours-R}
			& 0.0132 & 0.0162 & 0.0207& 0.0161 \\ \cline{2-7}
			&\multicolumn{2}{c|}{Ours-S}
			& 0.0130 & \pmb{0.0156} & \pmb{0.0199} & 0.0156 \\ \cline{2-7}
			\hline
		\end{tabular}
	\end{center}
\end{table}

\begin{figure*}[!tp]
	\centering
	\includegraphics[width=\textwidth]{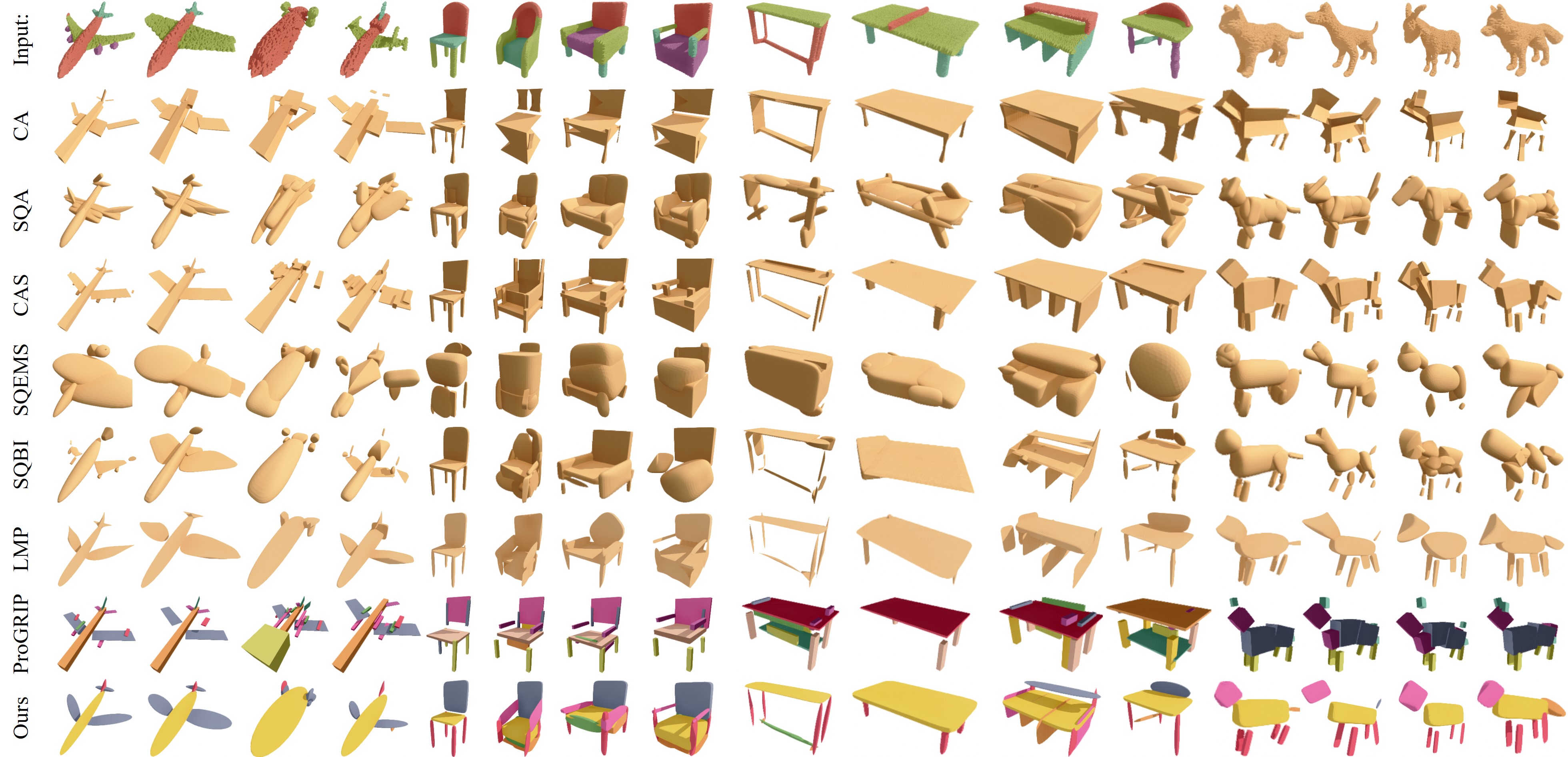}
	\caption{
		Comparison of semantic segmentation results. ProGRIP \cite{deng2022unsupervised} uses 4096 points with normal information in stage one and 4096 points in stage two. CoPartSeg \cite{umam2024unsupervised} uniformly samples 1024 points. Our approach distinguishes the primary components of objects across various categories and marks them with yellow coding.
	}
	\label{fig: shape abstraction}
\end{figure*}

Detailed quantitative results are presented in TABLE \ref{tab: shape abstraction}, while qualitative outcomes are illustrated in Fig. \ref{fig: shape abstraction}. CA, SQA, and CAS utilize point normal constraints to identify small-scale object parts. Consequently, ProGRIP benefit from CAS, facilitating the identification of animal ears, as depicted in Fig. \ref{fig: shape abstraction}. However, these normal constraints can lead to over-segmentation. Non-learning-based approaches like SQEMS and SQBI treat shape abstraction and segmentation as independent tasks performed sequentially, which hinders their ability to achieve performance comparable to CAS, LMP, and our method, where these tasks mutually enhance each other via joint optimization. Specifically, in terms of CD, our method performs similarly to LMP and significantly outperforms other methods. Regarding EMD, our method demonstrates notable superiority over LMP, particularly in the table category. Moreover, our approach surpasses LMP by generating semantic shape abstractions and identifying repeatable parts. 

\begin{table}[]
	\renewcommand\arraystretch{1.2}
	\begin{center}
		\caption{Quantitative comparison of hyperparameters. Semantic segmentation (Sem.) and instance segmentation (Ins.) }
		\label{tab: hyperparameter}
		\begin{tabular}{|c|cc|c|c|c|c|}
			\hline
			\multirow{2}{*}{Metric} & \multicolumn{2}{c|}{Hyperparameter} & \multirow{2}{*}{Airplane} & \multirow{2}{*}{Chair} & \multirow{2}{*}{Table} & \multirow{2}{*}{Animal} \\ \cline{2-3}
			& \multicolumn{1}{c|} {$S$} & $D$ & & & & 
			\\ \hline
			\multirow{9}{*}{\makecell[c]{mIoU$\uparrow$ \\ Ins.}}
			& \multicolumn{1}{c|}{\multirow{3}{*}{ \makecell[c]{6}}}
			& 128 & 0.5314 & 0.6291 & 0.5102 & - \\ \cline{3-7} 
			& \multicolumn{1}{c|}{}
			& 192 & 0.5756 & 0.6386 & 0.5399 & - \\ \cline{3-7} 
			& \multicolumn{1}{c|}{}
			& 256 & 0.6484 & 0.6363 & 0.5616 & - \\ \cline{2-7} 
			& \multicolumn{1}{c|}{\multirow{3}{*}{ \makecell[c]{8}}} 
			& 128 & 0.5199 & 0.4139 & 0.4307 & - \\ \cline{3-7} 
			& \multicolumn{1}{c|}{}
			& 192 & \pmb{0.6935} & 0.4901 & 0.4334 & - \\ \cline{3-7} 
			& \multicolumn{1}{c|}{}
			& 256 & 0.4595 & \pmb{0.6484} & 0.5346 & - \\ \cline{2-7} 
			& \multicolumn{1}{c|}{\multirow{3}{*}{ \makecell[c]{10}}} 
			& 128 & 0.6201 & 0.6459 & \pmb{0.5807} & - \\ \cline{3-7} 
			& \multicolumn{1}{c|}{}
			& 192 & 0.6638 & 0.6427 & 0.5705 & - \\ \cline{3-7} 
			& \multicolumn{1}{c|}{}
			& 256 & 0.4656 & 0.6079 & 0.4811 & - \\ \cline{3-7} 
			\hline
			\multirow{9}{*}{\makecell[c]{mIoU$\uparrow$ \\ Sem.}}
			& \multicolumn{1}{c|}{\multirow{3}{*}{ \makecell[c]{6}}}
			& 128 & \pmb{0.6869} & \pmb{0.2269} & \pmb{0.4858} & - \\ \cline{3-7} 
			& \multicolumn{1}{c|}{}
			& 192 & 0.6134 & 0.2069 & 0.4405 & - \\ \cline{3-7} 
			& \multicolumn{1}{c|}{}
			& 256 & 0.5838 & 0.1820 & 0.4066 & - \\ \cline{2-7} 
			& \multicolumn{1}{c|}{\multirow{3}{*}{ \makecell[c]{8}}} 
			& 128 & 0.5199 & 0.2139 & 0.4307 & - \\ \cline{3-7} 
			& \multicolumn{1}{c|}{}
			& 192 & 0.5335 & 0.1901 & 0.4334 & - \\ \cline{3-7} 
			& \multicolumn{1}{c|}{}
			& 256 & 0.5228 & 0.1745 & 0.3721 & - \\ \cline{2-7} 
			& \multicolumn{1}{c|}{\multirow{3}{*}{ \makecell[c]{10}}} 
			& 128 & 0.5108 & 0.1888 & 0.3976 & - \\ \cline{3-7} 
			& \multicolumn{1}{c|}{}
			& 192 & 0.4631 & 0.1683 & 0.3869 & - \\ \cline{3-7} 
			& \multicolumn{1}{c|}{}
			& 256 & 0.4276 & 0.1519 & 0.3502 & - \\ \cline{3-7} 
			\hline
			\multirow{9}{*}{CD$\downarrow$}
			& \multicolumn{1}{c|}{\multirow{3}{*}{ \makecell[c]{6}}}
			& 128 & 0.0010 & 0.0010 & 0.0011 & 0.0011 \\ \cline{3-7} 
			& \multicolumn{1}{c|}{}
			& 192 & \pmb{0.0006} & \pmb{0.0009} & 0.0011 & 0.0011 \\ \cline{3-7} 
			& \multicolumn{1}{c|}{}
			& 256 & 0.0006 & 0.0009 & 0.0013 & 0.0011 \\ \cline{2-7} 
			& \multicolumn{1}{c|}{\multirow{3}{*}{ \makecell[c]{8}}} 
			& 128 & 0.0007 & 0.0010 & 0.0011 & 0.0012 \\ \cline{3-7} 
			& \multicolumn{1}{c|}{}
			& 192 & 0.0006 & 0.0009 & 0.0011 & 0.0010 \\ \cline{3-7} 
			& \multicolumn{1}{c|}{}
			& 256 & 0.0007 & 0.0010 & 0.0013 & 0.0012 \\ \cline{2-7} 
			& \multicolumn{1}{c|}{\multirow{3}{*}{ \makecell[c]{10}}} 
			& 128 & 0.0008 & 0.0009 & 0.0010 & 0.0010 \\ \cline{3-7} 
			& \multicolumn{1}{c|}{}
			& 192 & 0.0006 & 0.0009 & 0.0010 & 0.0010\\ \cline{3-7} 
			& \multicolumn{1}{c|}{}
			& 256 & 0.0006 & 0.0009 & \pmb{0.0010} & \pmb{0.0010} \\ \cline{3-7} 
			\hline
		\end{tabular}
	\end{center}
\end{table}

\subsection{Hyperparameter}
\label{sec: membership_abl}
Our method necessitates the definition of several key hyperparameters: the number of primitives $M$, the number of semantics $S$, the number of points sampled on each primitive surface $I$, and the dimension of geometric features $D$, which is identical to the dimension of pose features. The values of $M$ and $I$ primarily influence instance-level shape abstraction and segmentation. Since our primary objective is to investigate the semantic representation of objects, we set $M=16$ and $I=256$, in accordance with the recommendations of LMP \cite{li2024shared}. To assess the impact of $S$ and $D$ on our method, we conduct experiments with $S$ values of $\{6, 8, 10\}$ and $D$ values of $\{128, 192, 256\}$. Detailed quantitative comparison results for instance segmentation and semantic segmentation based on mIoU, and shape abstraction with repeatable primitives based on CD are presented in TABLE \ref{tab: hyperparameter}. Intuitively, a larger $S$ enhances the capability of our model to handle semantically complex man-made objects. However, as shown in TABLE \ref{tab: hyperparameter}, a larger $S$ also increases the likelihood of misclassifying parts with the same semantics into different semantics due to minor errors. Concurrently, a larger $D$ improves the geometric adaptability of primitives, as evidenced by the CD values in TABLE \ref{tab: hyperparameter}, but it can lead to overfitting, causing the semantic representation to degrade into instance representation. Taking these factors into account, we choose $S=6$ and $D=192$ to achieve a balance between semantic and instance representation.

\begin{table}
	\renewcommand\arraystretch{1.2}
	\begin{center}
		\caption{Summary of ablation study. The temperature parameter $\tau$ is marked as adaptive if checked, otherwise it is fixed at 1. Sparsity indicates whether Sparsemax or Softmax is applied.}
		\label{tab: ablation}
		\begin{tabular}{| c | c | c | c | c | c | c | c |}
			\hline
			Case & $\mathcal{L}_{\text{recon}} $ & $\mathcal{L}_{\text{hd}} $ & $\mathcal{L}_{\text{wd}} $ & $\mathcal{L}_{\text{c}} $ & $\mathcal{L}_{\text{a}} $ & $\tau$ & Sparsity
			\\ \hline
			({\romannumeral1})
			& \checkmark & \usym{2717} &\usym{2717} & \usym{2717} & \usym{2717} & \checkmark & \checkmark \\ \hline
			({\romannumeral2})
			& \checkmark & \checkmark & \usym{2717}& \usym{2717} & \usym{2717} & \checkmark & \checkmark \\ \hline
			({\romannumeral3})
			& \checkmark & \checkmark &\checkmark & \usym{2717} & \usym{2717} & \checkmark & \checkmark \\ \hline
			({\romannumeral4})
			& \checkmark & \checkmark &\usym{2717} & \checkmark & \usym{2717} & \checkmark & \checkmark \\ \hline
			({\romannumeral5})
			& \checkmark & \checkmark &\usym{2717} & \usym{2717} & \checkmark & \checkmark & \checkmark \\ \hline
			({\romannumeral6})
			& \checkmark & \checkmark &\checkmark & \checkmark & \usym{2717} & \checkmark & \checkmark \\ \hline
			({\romannumeral7})
			& \checkmark & \checkmark &\checkmark & \usym{2717} & \checkmark & \checkmark & \checkmark \\ \hline
			({\romannumeral8})
			& \checkmark & \checkmark &\usym{2717} & \checkmark & \checkmark & \checkmark & \checkmark \\ \hline
			({\romannumeral9})
			& \checkmark & \checkmark &\checkmark & \checkmark & \checkmark & \usym{2717} & \checkmark \\ \hline
			({\romannumeral10})
			& \checkmark & \checkmark &\checkmark & \checkmark & \checkmark & \checkmark & \usym{2717} \\ \hline
			({\romannumeral11})
			& \checkmark & \checkmark &\checkmark & \checkmark & \checkmark & \checkmark & \checkmark \\ \hline
		\end{tabular}
	\end{center}
\end{table}

\begin{figure}[!tbp]
	\centering
	\includegraphics[width=\linewidth]{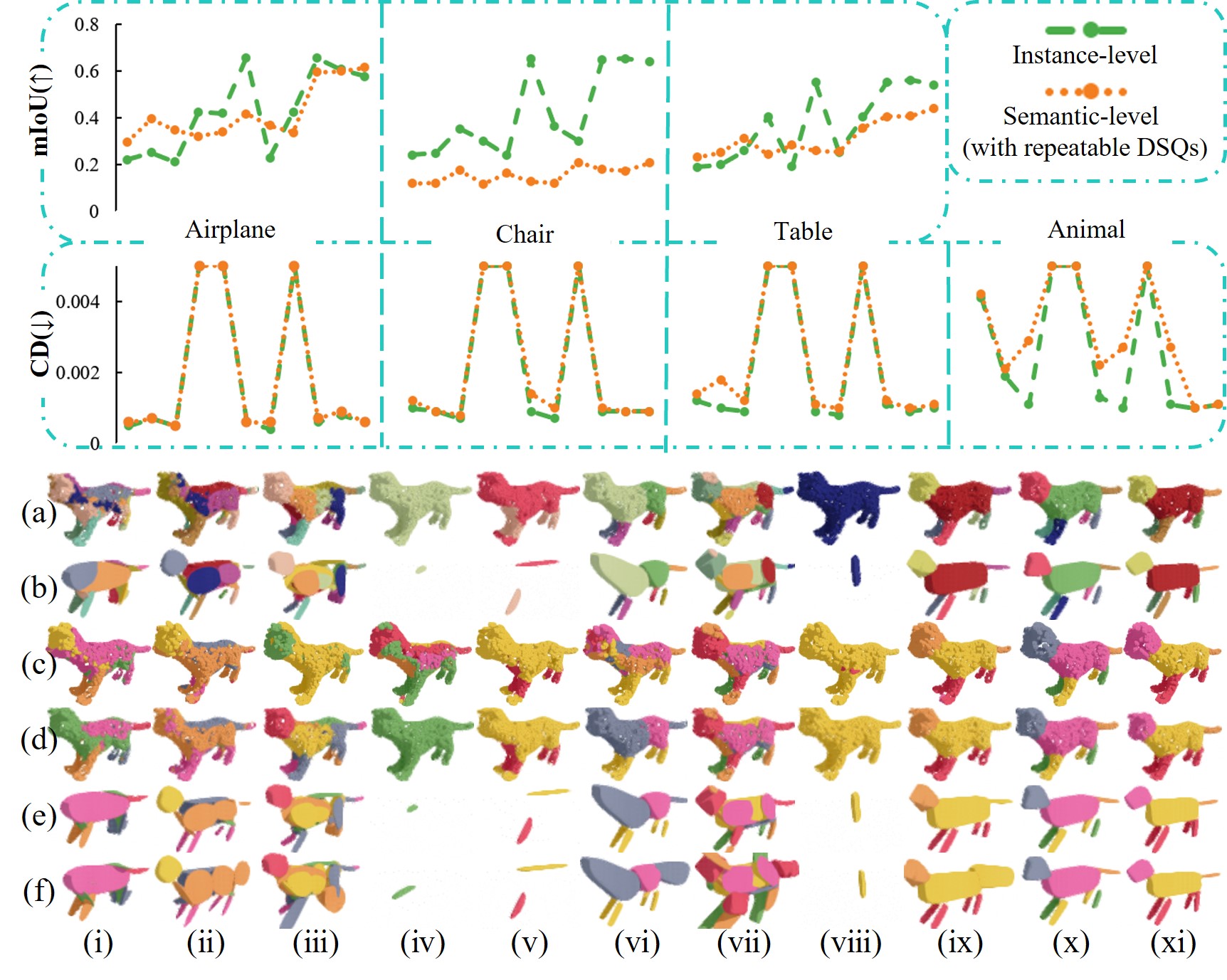}
	\caption{Quantitative and qualitative ablation experiments. mIoU for instance / semantic segmentation, and CD for instance shape abstraction and shape abstraction with repeatable primitives (DSQs). Each line plot in the graph presents data from 11 ablation cases listed in Table \ref{tab: ablation}. Note that the CD values for cases ({\romannumeral4}), ({\romannumeral5}), and ({\romannumeral8}) were capped at a maximum of 0.005 for clarity. The visualized cat samples, arranged from top to bottom, include the following: (a) instance segmentation, (b) instance shape abstraction, (c) segmentation based on alignment similarity, (d) semantic segmentation, (e) semantic shape abstraction, and (f) shape abstraction with repeatable primitives. (a) and (b) share the same color-codes, while the others share different color-codes.}
	\label{fig: ablation}
\end{figure}

\subsection{Ablation study}
To evaluate the significance of each loss function, the adaptive temperature $\tau$ in (\ref{eq: tau}), and Sparsemax, we conduct a comparison among 11 training scenarios, all sharing identical hyperparameters except for variations detailed in TABLE \ref{tab: ablation}. Clarification is warranted that Case (\romannumeral9) enforces $\tau$ to be constant at 1, while Case (\romannumeral10) replaces Sparsemax in (\ref{eq: slmp matrixs}) with Softmax. The comparison results are depicted in Fig. \ref{fig: ablation}. Compared to animals, the shapes of airplanes, chairs, and tables are more rigid. As a result, $\mathcal{L}_{\text{hd}}$ shows more pronounced effects in advancing shape reconstruction within animals. Additionally, in the presence of constraints such as $\mathcal{L}_{\text{c}}$ or $\mathcal{L}_{\text{a}}$, the model may converge to trivial solutions if $\mathcal{L}_{\text{wd}}$ is not applied to maintain the instance-level and semantic-level membership. $\mathcal{L}_{\text{c}}$ tends to produce singular instances and semantic trivial solutions more readily than $\mathcal{L}_{\text{a}}$. Comparing ({\romannumeral6}) and ({\romannumeral11}), as well as ({\romannumeral9}) and ({\romannumeral10}), $\mathcal{L}_{\text{a}}$ is observed to regulate the alignment of instances and semantics. Examining (f) in ({\romannumeral9}) and ({\romannumeral11}), the MSE-based adaptive $\tau$ expands semantic differences within individual shapes to mitigate semantic ambiguity. Furthermore, comparing ({\romannumeral10}) and ({\romannumeral11}), Sparsemax reduces semantic interference among shared geometric feature spaces through sparsity. Specifically, replacing Sparsemax with Softmax results in the learned geometric features of object parts tending to distribute more evenly across the entire geometric feature space, rather than being confined to easily aligned subspaces, as depicted in Fig. \ref{fig: keypoint}.

\begin{figure}[!tbp]
	\centering
	\includegraphics[width=\linewidth]{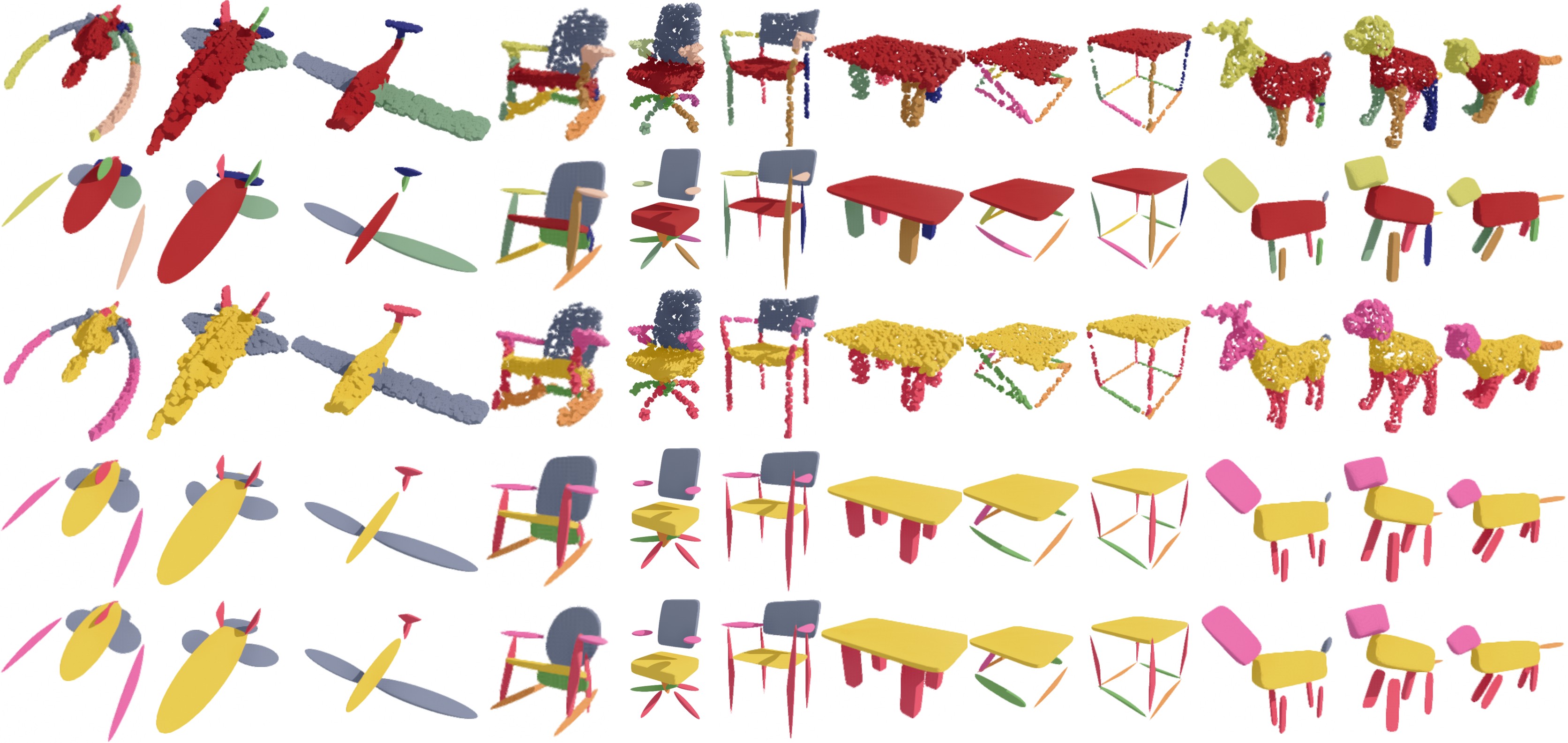}
	\caption{Visualization of our shape abstraction and segmentation results across the four object categories. 1st row: instance segmentation; 2nd row: instance shape abstraction; 3rd row: semantic segmentation; 4th row: semantic shape abstraction; 5th row: shape abstraction with repeatable primitives. The 1st and 2nd rows shares the same color-codes, whereas the 3rd, 4th, and 5th shares the same color-codes.}
	\label{fig: allresults}
\end{figure}

To provide a clearer comparison, Fig. \ref{fig: allresults} illustrates the five final results generated by our fully unsupervised, one-stage, joint semantic and instance learning-based 3D representation model, without requiring any supervision or heuristic information.

\subsection{Encoder Comparison}
\begin{figure}[!tbp]
	\centering
	\includegraphics[width=\linewidth]{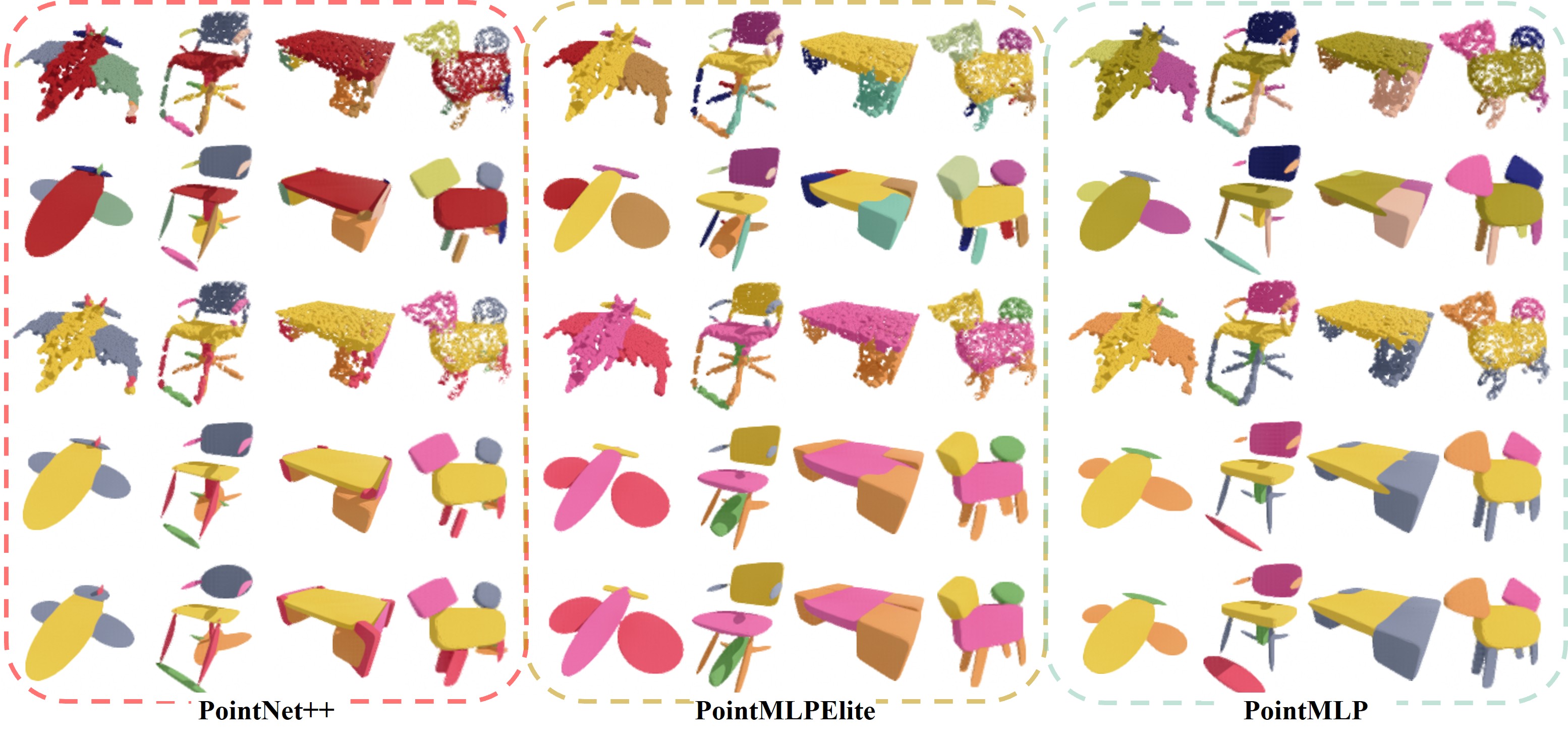}
	\caption{
		Visualization of shape abstraction and segmentation results across different encoders. The first and second rows (instance segmentation and instance shape abstraction) share a common color scheme, while the third, fourth, and fifth rows (semantic segmentation, semantic shape abstraction, and shape abstraction with repeatable primitives) share the same color-codes. Note that different encoders employ distinct color schemes.
	}
	\label{fig: encoder comparison}
\end{figure}

To examine the influence of different encoders on model performance, we conducted experiments with 3D point cloud backbones, specifically PointNet++ \cite{qi2017pointnet++}, PointMLP \cite{ma2022rethinking}, and its variant PointMLPElite \cite{ma2022rethinking}. Given that transformer-based models \cite{zhao2021point, wu2022point, wu2024point} are overly complex for our task, we focused on the aforementioned backbones: PointNet++ (1.9M parameters), PointMLPElite (1.4M parameters), and PointMLP (16.8M parameters). The visualization results are provided in Fig. \ref{fig: encoder comparison}, while the quantitative results are summarized in Table \ref{tab: encoder comparison}.

\begin{table}
	\renewcommand\arraystretch{1.2}
	\begin{center}
		\caption{Quantitative Comparison of Segmentation and Shape Abstraction Performance Across Different Encoders: mean Intersection over Union (mIoU) for Instance Segmentation (Ins. S) and Semantic Segmentation (Sem. S), and Chamfer Distance (CD-$ \ell^2 $) for Instance Shape Abstraction (Ins. SA), Semantic Shape Abstraction (Sem. SA) and Shape Abstraction with Repeatable DSQs (Rep. SA).}
		\label{tab: encoder comparison}
		\begin{tabular}{|c|cp{0.1cm}|c|c|c|c|}
			\hline
			Metric & \multicolumn{2}{c|}{Encoder} & Airplane & Chair & Table & Animal \\ \hline
			\multirow{3}{*}{\makecell[c]{mIoU$\uparrow$ \\ Ins. S}}
			& \multicolumn{2}{c|}{PointNet++} 
			& 0.5189 & 0.5169 & 0.4462 & -\\ \cline{2-7}
			& \multicolumn{2}{c|}{PointMLPElite}
			& \pmb{0.5827} & \pmb{0.6461} & \pmb{0.5381} & -\\ \cline{2-7}
			&\multicolumn{2}{c|}{PointMLP} 
			& 0.5662 & 0.6350 & 0.5223 & -\\ \cline{2-7}
			\hline
			\multirow{3}{*}{\makecell[c]{mIoU$\uparrow$ \\ Sem. S}}
			& \multicolumn{2}{c|}{PointNet++} 
			& \pmb{0.6134} & 0.2069 & 0.4405 & -\\ \cline{2-7}
			& \multicolumn{2}{c|}{PointMLPElite}
			& 0.6060 & 0.1998 & \pmb{0.5294} & -\\ \cline{2-7}
			&\multicolumn{2}{c|}{PointMLP} 
			& 0.5828 & \pmb{0.2079} &0.4861 & -\\ \cline{2-7}
			\hline
			\multirow{3}{*}{\makecell[c]{CD$\downarrow$ \\ Ins. SA}}
			& \multicolumn{2}{c|}{PointNet++} 
			& \pmb{0.0006} & \pmb{0.0009} & \pmb{0.0010} & \pmb{0.0011} \\ \cline{2-7}
			& \multicolumn{2}{c|}{PointMLPElite}
			& 0.0008 & 0.0011 & 0.0014 & 0.0013 \\ \cline{2-7}
			&\multicolumn{2}{c|}{PointMLP} 
			& 0.0007 & 0.0011 & 0.0011 &\pmb{0.0011} \\ \cline{2-7}
			\hline
			\multirow{3}{*}{\makecell[c]{CD$\downarrow$ \\ Sem. SA}}
			& \multicolumn{2}{c|}{PointNet++} 
			& \pmb{0.0006} & \pmb{0.0009} & \pmb{0.0011} & \pmb{0.0011} \\ \cline{2-7}
			& \multicolumn{2}{c|}{PointMLPElite}
			& 0.0008 & 0.0011 & 0.0014 & 0.0013\\ \cline{2-7}
			&\multicolumn{2}{c|}{PointMLP} 
			& 0.0007 & 0.0011 & 0.0011 & \pmb{0.0011}\\ \cline{2-7}
			\hline
			\multirow{3}{*}{\makecell[c]{CD$\downarrow$ \\ Rep. SA}}
			& \multicolumn{2}{c|}{PointNet++} 
			& \pmb{0.0006} & \pmb{0.0009} & \pmb{0.0010} & \pmb{0.0011} \\ \cline{2-7}
			& \multicolumn{2}{c|}{PointMLPElite}
			& 0.0009 & 0.0012 & 0.0015 & 0.0014 \\ \cline{2-7}
			&\multicolumn{2}{c|}{PointMLP} 
			& 0.0007 & 0.0012 & 0.0011 & 0.0012 \\ \cline{2-7}
			\hline
		\end{tabular}
	\end{center}
\end{table}

As indicated in Table \ref{tab: encoder comparison} and Fig. \ref{fig: encoder comparison}, more advanced encoders enhance instance segmentation performance but have minimal impact on semantic segmentation and shape abstraction. Specifically, for instance segmentation, more advanced encoders are better at capturing finer geometric details, such as the contours of airplane wings and the legs of chairs and tables in Fig. \ref{fig: encoder comparison}. However, for semantic segmentation, the sparse representation of object part geometric features we propose encourages semantically similar features to align in the same low-dimensional subspace, emphasizing shared abstract geometric features over fine-grained details. Therefore, the effect of more advanced encoders on semantic segmentation is minimal. Notably, PointMLPElite excels in semantic segmentation for tables, likely attributable to its simplicity, which is particularly effective for categories with less shape variation, as illustrated in Fig. \ref{fig: encoder comparison}. Regarding shape abstraction, PointMLP does not show a significantly improved reconstruction of the input point cloud within the same constrained superquadric parameter space, compared to PointNet++.

\subsection{Results on Other Categories}
\begin{figure}[!tbp]
	\centering
	\includegraphics[width=\linewidth]{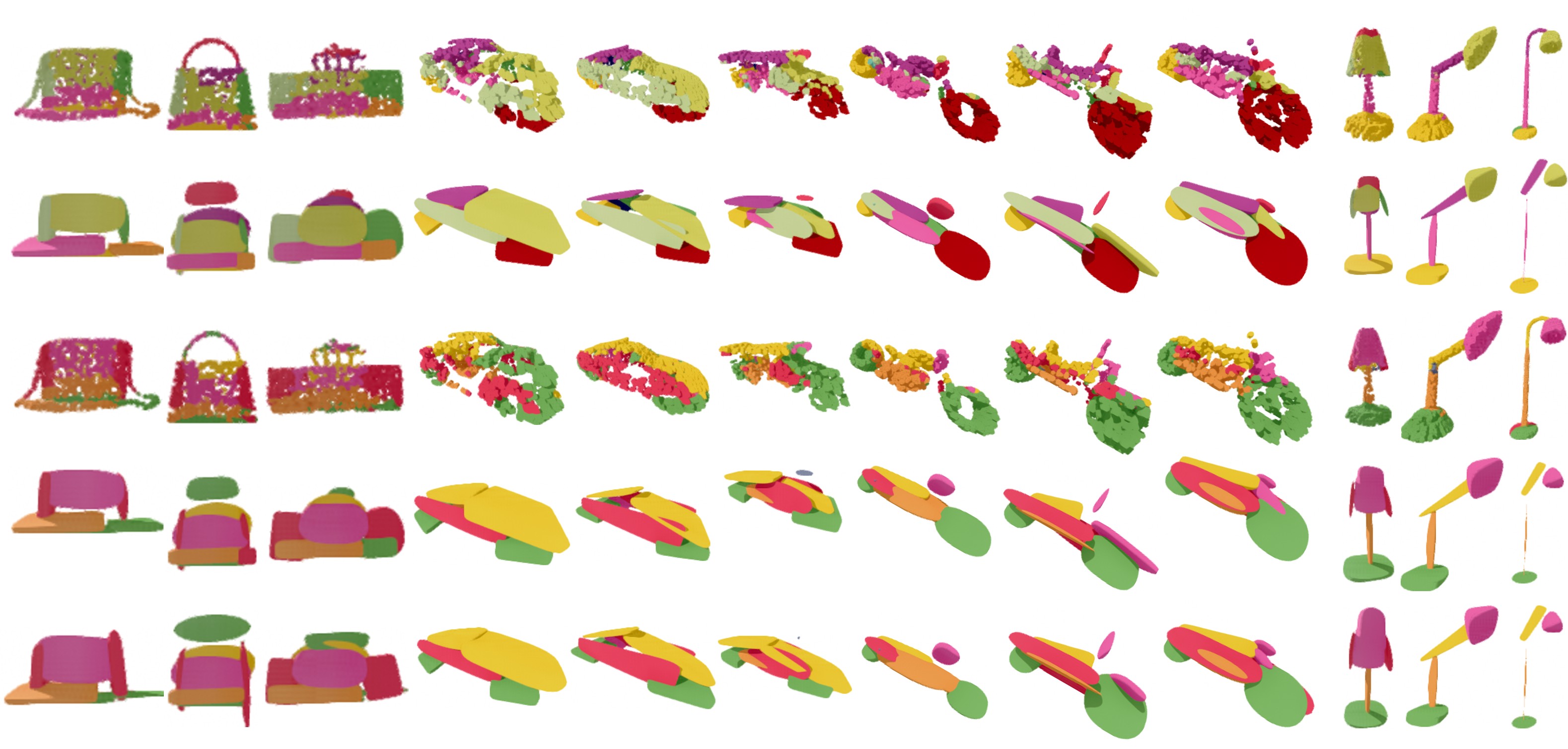}
	\caption{Visualization of shape abstraction and segmentation results across various categories: bags, cars, motorbikes, and lamps. 1st row: instance segmentation; 2nd row: instance shape abstraction; 3rd row: semantic segmentation; 4th row: semantic shape abstraction; 5th row: shape abstraction with repeatable primitives. The 1st and 2nd rows shares the same color-codes, whereas the 3rd, 4th, and 5th shares the same color-codes.}
	\label{fig: more shapes}
\end{figure}

Beyond the initial four categories, we expand our analysis to include four additional categories from ShapeNetCore \cite{yi2016scalable}: bags, cars, motorbikes, and lamps, as depicted in Fig. \ref{fig: more shapes}. Despite the inherent complexity of these shapes, which are not well-suited for superquadric-based shape abstraction—such as the handles of bags, the ambiguous contours of car wheels, the unclear geometry of motorbikes, and the irregular surfaces of lamps—our method successfully identifies potential repeatable parts, including the wheels of both cars and motorbikes.

\section{CONCLUSIONS}
This paper presents a fully unsupervised, one-stage method that leverages sparse representation and feature alignment to unify semantic- and instance-level shape representations from unoriented object point clouds, producing outputs including instance segmentation, semantic segmentation, instance shape abstraction, semantic shape abstraction, and shape abstraction with repeatable primitives. For sparse representation, we introduce the Sparsemax function to construct instance-level part pose features and both instance-level and semantic-level part geometric features in a high-dimensional space, where each object part feature is represented as a sparse convex combination of point features. In this way, semantically similar part features lie in or near the same subspaces. Feature alignment involves a customized attention-based strategy to align instance-level part geometric features and semantic-level part geometric features and reconstructing the input point cloud using pairs of instance-level pose features with instance-level geometric features and semantic-level geometric features. Our approach operates without the need for specific training for each object category and does not rely on any supervision or heuristic information. Our predefined number of semantics is sufficient for most shape abstraction tasks, which highlight the key geometry of object shapes. Experimental results demonstrate that, through the sparsity and alignment, our approach effectively unveils the semantics of object shapes.


\end{document}